\documentclass[lettersize,journal]{IEEEtran}
\usepackage{cite}
\usepackage{amsmath,amssymb,amsfonts}%
\usepackage{algorithmic}
\usepackage{array}
\usepackage[caption=false,font=normalsize,labelfont=sf,textfont=sf]{subfig}
\usepackage{textcomp}
\usepackage{stfloats}
\usepackage{url}
\usepackage{verbatim}
\usepackage{graphicx}

\usepackage{multirow}%

\usepackage{amsthm}%

\usepackage{xcolor}%
\usepackage{textcomp}%
\usepackage{manyfoot}%
\usepackage{booktabs}%
\usepackage{listings}%
\usepackage{pifont, comment, hhline, float, placeins, array, bm, makecell, dsfont, multicol, caption, anyfontsize}
\def\BibTeX{{\rm B\kern-.05em{\sc i\kern-.025em b}\kern-.08em
    T\kern-.1667em\lower.7ex\hbox{E}\kern-.125emX}}
\usepackage{balance}
\begin{document}
\title{Exploiting the English Grammar Profile for L2 grammatical analysis with LLMs\thanks{This work has been submitted to the IEEE for possible publication. Copyright may be transferred without notice, after which this version may no longer be accessible.}}
\author{Stefano Bannò, Penny Karanasou, Kate Knill, \IEEEmembership{Senior Member, IEEE}, Mark Gales, \IEEEmembership{Fellow, IEEE}\thanks{The authors are with the Institute for Automated Language Teaching and Assessment and the Machine Intelligence Laboratory of the Cambridge University Engineering Department, Trumpington Street, Cambridge CB2 1PZ, UK. Stefano Bannò is the corresponding author (e-mail: sb2549@cam.ac.uk).}}

\markboth{Journal of \LaTeX\ Class Files,~Vol.~18, No.~9, March~2026}%
{How to Use the IEEEtran \LaTeX \ Templates}

\maketitle

\begin{abstract}

Evaluating the grammatical competence of second language (L2) learners is essential both for providing targeted feedback and for assessing proficiency. To achieve this, we propose a novel framework leveraging the English Grammar Profile (EGP), a taxonomy of grammatical constructs mapped to the proficiency levels of the Common European Framework of Reference (CEFR), to detect learners’ attempts at grammatical constructs and classify them as successful or unsuccessful. This detection can then be used to provide fine-grained feedback. Moreover, the grammatical constructs are used as predictors of proficiency assessment by using automatically detected attempts as predictors of holistic CEFR proficiency. For the selection of grammatical constructs derived from the EGP, rule-based and LLM-based classifiers are compared. We show that LLMs outperform rule-based methods on semantically and pragmatically nuanced constructs, while rule-based approaches remain competitive for constructs that rely purely on morphological or syntactic features and do not require semantic interpretation. For proficiency assessment,  we evaluate both rule-based and hybrid pipelines and show that a hybrid approach combining a rule-based pre-filter with an LLM consistently yields the strongest performance. Since our framework operates on pairs of original learner sentences and their corrected counterparts, we also evaluate a fully automated pipeline using automatic grammatical error correction. This pipeline closely approaches the performance of semi-automated systems based on manual corrections, particularly for the detection of successful attempts at grammatical constructs. Overall, our framework emphasises learners’ successful attempts in addition to unsuccessful ones, enabling positive, formative feedback and providing actionable insights into grammatical development.
\end{abstract}

\begin{IEEEkeywords}
Second language assessment, automatic writing evaluation, grammatical feedback, computer-assisted language learning.
\end{IEEEkeywords}

\section{Introduction}
\label{sec:intro}
\IEEEPARstart{A}{utomated} essay scoring (AES) has become a well-established research domain within natural language processing (NLP), significantly contributing to language education and computer-assisted language learning (CALL)~\cite{klebanov2022automated, li2024essay}. A central goal in this field is the assessment of second language (L2) proficiency, with growing emphasis not only on holistic scoring but also on analysing specific aspects of language ability to support more targeted and pedagogically valuable feedback~\cite{hamplyons1995rating,weigle2002assessing}. Among the various dimensions of language proficiency, grammar plays a key role in L2 learning, enabling learners to structure meaning and communicate with clarity and precision. It is ``the weaving that creates the fabric'' of language, providing the structural cohesion that binds vocabulary and meaning into coherent expression~\cite[p. 2]{azar2007grammar}.

Within this broader context, grammatical error detection (GED) and correction (GEC) have become key areas in the field of CALL~\cite{bryant2023grammatical}. In addition to their standalone value, features derived from grammatical errors have also been found effective for AES~\cite{yannakoudakis2011, gamon} since errors can serve as meaningful predictors of proficiency~\cite{hawkins2010}.
However, while GED and GEC aim to improve learners' grammatical \emph{accuracy}, they often do not capture the nuances of learners' grammatical \emph{complexity}.
Far from being a monolithic and clearly defined concept~\cite{bulte2012defining}, grammatical complexity has been predominantly investigated by giving priority to its clausal, syntactic level~\cite{lan2019grammaticalcomplexity}, which is also evident in numerous works that use mainly syntactic features for writing~\cite{wolfe1998second, ortega2003syntactic,lu2010automatic,mcnamara2010linguistic} and speaking~\cite{iwashita2006syntactic, chen2011computing, yoon2012assessment} assessment. Moreover, many of these studies focus on complexity indices that, while effective predictors of L2 proficiency, are not pedagogically actionable and thus offer limited guidance for classroom instruction or learner development.
Our work addresses these limitations by focusing on the automated analysis of L2 grammatical complexity -- broadly defined beyond syntactic structures --  to support both assessment and fine-grained feedback.

We
leverage the English Grammar Profile 
(EGP)~\cite{okeefe2017english} (see Section \ref{sec:egp}), a systematically validated and evidence-based taxonomy of English grammar that categorises grammatical constructs in the form of clearly interpretable \emph{can-do} statements, describing what a learner can do at different proficiency levels of the Common European Framework of Reference (CEFR) \cite{cefr2001}. A grammatical construct refers to a specific arrangement or pattern of words and phrases that follows the rules of grammar in a language and serves a particular function in communication. In the EGP, among other aspects, a grammatical construct can include:
\begin{itemize}
\item [a.] tense structures (e.g., past simple passive: \emph{The documents were signed by the manager.})
\item [b.]  conditional clauses (e.g., \emph{If I had known, I would have called.})
\item [c.]  use of adjectives (e.g., \emph{She is seeing another person.})

\item[d.] pragmatic appropriateness (e.g., use of politeness forms: \emph{I wondered if you could introduce me to somebody.})\footnote{As in GED and GEC, the definition of \emph{grammatical construct}, and in particular, the use of the term \emph{grammatical}, is applied here in a broad sense. A grammatical construct may encompass morphological (e.g., example \emph{a}), syntactic (e.g., example \emph{b}), semantic (e.g., example \emph{c}), and pragmatic (e.g., example \emph{d}) aspects.}
\end{itemize}
Consistent with the examples above, representative \emph{can-do} statements from the EGP include: \emph{Can use the past simple passive affirmative with a range of pronoun and noun subjects} (B1 level); \emph{Can use a finite subordinate clause, before or after a main clause, with conjunctions to introduce conditions} (B2 level); \emph{Can use `another' to talk about something different} (B1 level); and \emph{Can use the past simple with `I wondered' and `I wanted' as politeness structures, when making polite requests and thanking} (B2 level). This resource adds significant depth to the study of L2 grammar by highlighting granular elements of grammatical complexity and allowing for a more extensive understanding of L2 development beyond grammatical accuracy and traditional measures of syntactic and grammatical complexity.

\begin{figure}[htbp!]
    \centering
    \includegraphics[width=0.7\linewidth]{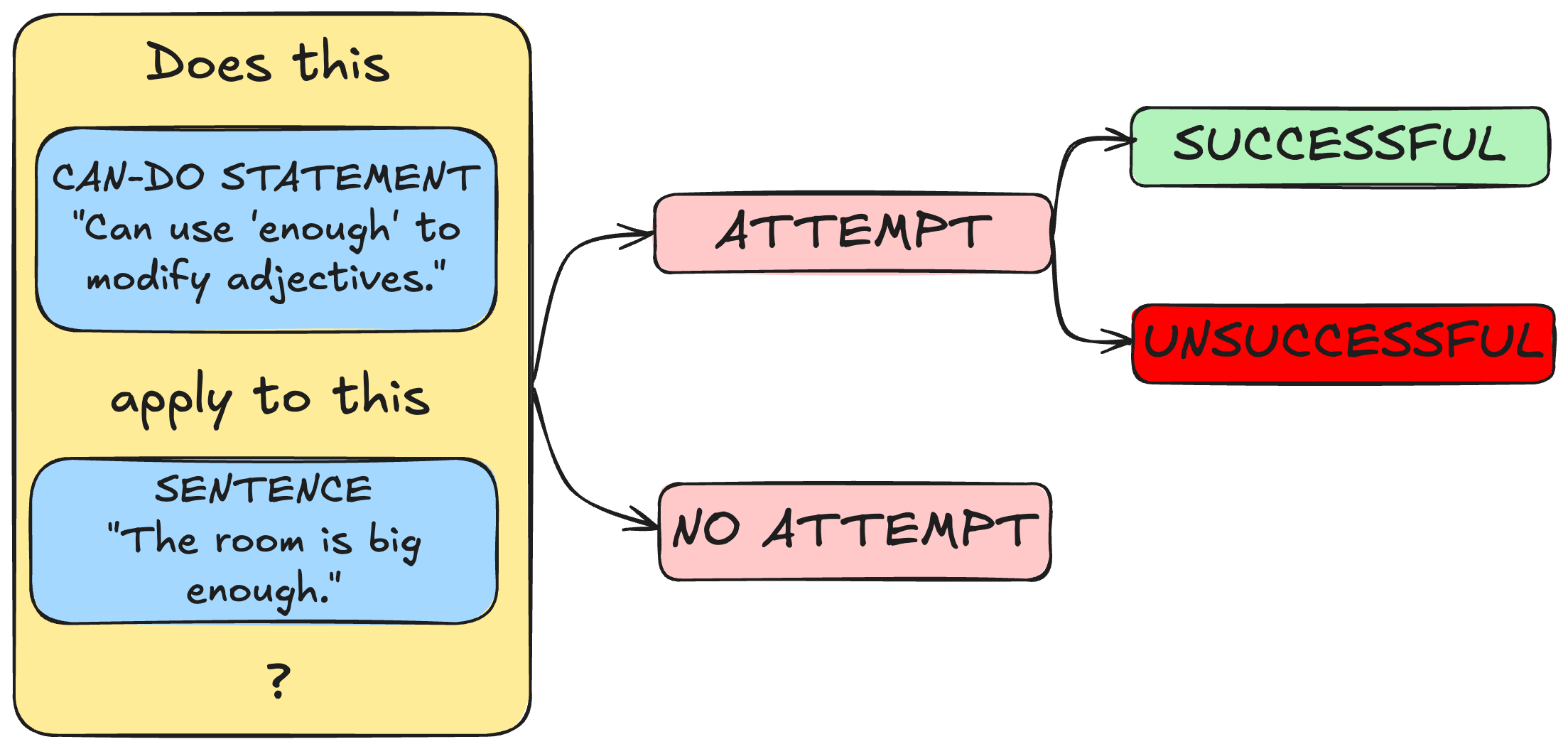}
    \caption{Diagram of the proposed attempt-based framework.}
    \label{fig:egp_diagram}
\end{figure}

Understanding what a learner is attempting to express when writing is fundamental for teachers and testers in order to better provide feedback. However, reliably detecting learners' attempts at specific grammatical constructs -- and determining whether those attempts are successful or not -- remains a significant challenge. We propose to apply the \emph{can-do} statements drawn from the EGP to both learner-produced sentences and their grammatically corrected (manually or automatically) counterparts. By comparing the outcome in both versions of each sentence, we aim to develop a framework, and a related task, for identifying whether grammatical constructs are attempted or not, and if attempted, whether they are successfully or unsuccessfully realised by L2 learners of English (see Figure \ref{fig:egp_diagram} and Section \ref{sec:grammatical_attempt} for more details). Beyond detecting such attempts, we also investigate their relationship with overall writing proficiency.

To exploit the information contained in the EGP, a model is needed that can accurately interpret L2 learner texts and detect the use of grammatical constructs linked to their related \emph{can-do} statements. Large language models (LLMs) seem to be a fitting choice for this task, as they have previously demonstrated the ability to interpret natural language descriptors for L2 holistic~\cite{yancey-etal-2023-rating} and analytic assessment~\cite{banno-etal-2024-gpt} and have been employed for analytic feedback of discourse coherence~\cite{naismith-etal-2023-automated} and vocabulary~\cite{banno-etal-2025-vocabulary}. Additionally, they have been extensively investigated for GEC~\cite{katinskaia-yangarber-2024-gpt, omelianchuk-etal-2024-pillars} and to explain grammatical errors~\cite{song-etal-2024-gee}. 
However, unlike GED and GEC, which focus primarily on errors (i.e., negative feedback), our work also emphasises what learners \emph{can} do, aligning with recent research that highlights the greater effectiveness of positive feedback \cite{kerr2020feedback}.

In our work, we address the following research questions (RQs):

\begin{itemize}
    \item [RQ1:] How can learners’ attempts at grammatical constructs be identified automatically and classified as successful or unsuccessful?
    \item [RQ2:] How can this information be leveraged to provide feedback?
    \item [RQ3:] How can this information be used for proficiency assessment?
\end{itemize}

Our contribution is threefold:

\begin{itemize}
\item [a.] We propose a novel framework for identifying learners’ attempts at grammatical constructs, as defined by the EGP, and determining whether these attempts are successful or unsuccessful (RQ1). The framework can be leveraged to provide grammatical feedback (RQ2).

\item [b.] We create a new annotated dataset, consisting of a subset of the L2 learner essays later used for proficiency assessment (see below), with a selection of 12 EGP \emph{can-do} statements as annotation targets.
We evaluate two LLMs on the task of identifying and classifying learners’ attempts at specific grammatical constructs (RQ1) and compare their performance with two rule-based systems, one of which is publicly available. On average, the LLMs outperform the rule-based baselines; however, for grammatical constructs defined purely by syntactic or morphological patterns, rule-based approaches remain more effective.

\item [c.] We show that predicted attempts at grammatical constructs, particularly successful ones, are effective predictors of writing proficiency, using a publicly available corpus of L2 learner essays (RQ3). In this assessment task, we compare a publicly available rule-based system with a hybrid pipeline combining rules and an LLM in a zero-shot setting. Results show that incorporating the LLM substantially improves performance on holistic scoring.

\end{itemize}

Our attempt-based framework uses pairs of original and corrected sentences. To support the above, a semi-automated pipeline, based on manual corrections, and a fully automated pipeline, using a GEC system to generate corrections, are introduced.

\section{Related work}
\label{sec:related}
\subsection{Grammatical complexity}
Grammatical proficiency is not only determined by grammatical accuracy but also by grammatical complexity. However, despite its widespread use in L2 research, grammatical complexity remains a contested construct, with no agreed-upon definition~\cite{lan2019grammaticalcomplexity} or consistent operationalisation across studies~\cite{ehret2023measuring}. Early definitions tended to rely on broad notions such as the use of gradually ``more elaborate language''~\cite[p. 303]{foster1996} or ``grammatical variation and sophistication''~\cite[p. 69]{wolfe1998second}, which lacked clear criteria. More recent accounts have proposed more structured taxonomies, distinguishing between systemic complexity (the range of grammatical forms a learner can produce) and structural complexity (the depth or sophistication of those forms)~\cite{bulte2012defining}. Others, particularly from register studies, argue that grammatical complexity is context-dependent and should be analysed in relation to communicative function~\cite{biber2011}. Recent work \cite{bulte2025complexity} calls for a narrow, standardised set of ratio-based complexity measures to enhance replicability, though this has sparked debate. Critics like \cite{biber2025grammatical} argue such metrics overlook syntactic functions, advocating instead for more fine-grained, functionally sensitive indices. 

Furthermore, over time, the focus of the studies on complexity has shifted from clausal features alone~\cite{wolfe1998second, ortega2003syntactic} to also include phrasal and morphological structures~\cite{biber2011, staples2016academic}, further highlighting the multidimensional nature of the construct~\cite{lan2019grammaticalcomplexity}.

We believe that the English Grammar Profile \cite{okeefe2017english}, which underpins our work, represents a major development in this multidimensional view of grammatical complexity. It includes over 1,000 grammatical constructs, each expressed through easily interpretable natural language \emph{can-do} statements, spanning clausal, morphological, and phrasal dimensions, and also including semantic and pragmatic (see examples \emph{c} and \emph{d} in Section \ref{sec:intro}, respectively) aspects of language. Each construct is associated with a CEFR proficiency level and is accompanied by real learner examples along with their grammatically corrected versions, which offer additional contextual information, and other descriptive metadata. Further details on this resource are provided in Section \ref{sec:egp}. Two other comparable resources are Pearson's Global Scale of English (GSE)\footnote{\url{english.com/gse/teacher-toolkit/user/grammar}} and the CEFR-J Grammar Profile.\footnote{\url{cefr-j.org/download.html\#cefrj_grammar}} We chose the EGP because it is available in a downloadable Excel format (unlike the GSE), is based on data from a wide range of first language (L1) backgrounds and includes learner examples (unlike the CEFR-J Grammar Profile).

\subsection{Automated approaches for assessment}

Trends in the theoretical literature are mirrored in both automated and semi-automated approaches to grammatical complexity analysis for L2 assessment and feedback. \cite{lu2010automatic} conducted a systematic study using an automated system, applying 14 syntactic measures to a large dataset of Chinese learners' writing samples. Their findings indicated that these syntactic measures were strong predictors of students' writing proficiency. \cite{mcnamara2010linguistic} investigated various linguistic features, including lexical, syntactic, and cohesion-related indices, and found syntactic complexity measures to be the most predictive of writing quality. However, their analysis focused exclusively on syntactic rather than broader grammatical features.

In line with shifts in the theoretical literature, \cite{biber2016predicting} moved beyond purely clausal syntactic features by incorporating morphological and phrasal elements in their analysis of grammatical complexity in TOEFL iBT spoken and written responses. Using part-of-speech (PoS) tagging and syntactic parsing, they identified variation in feature distribution across modality, task type, and proficiency level. Similarly, \cite{lu2017automated} highlighted the importance of complexity not only in L2 writing research but also in rating scales and AES systems, calling attention to mismatches between theoretical definitions and operational practices in assessment contexts. Neither study, however, implemented a fully automated process for feature extraction. Using a fully automated feature extraction approach based on the Tool for the Automatic Analysis of Syntactic Sophistication and Complexity (TAASSC)~\cite{kyle2016measuring}, \cite{kyle2018measuring} demonstrated that fine-grained indices of phrasal complexity were stronger predictors of L2 writing proficiency than traditional syntactic complexity measures.


\subsection{Automated approaches for feedback}
\label{sec:auto_approaches_feedback}

For clarity, we distinguish between two strands of research. The first, discussed in the previous paragraph, involves the development and exploration of grammatical complexity indices and measures which, while effective predictors of L2 proficiency, are often not easily interpretable or actionable for teachers. The second, which we focus on in this paragraph, includes studies that employ techniques designed to be more interpretable and usable in educational settings for fine-grained feedback. An early example of this line of work is CTAP (Common Text Analysis Platform), a web-based system introduced by \cite{chen-meurers-2016-ctap}, which provides extraction and visualisation of lexical, syntactic, and morpho-syntactic complexity features, aiming to bridge the gap between state-of-the-art NLP analyses and pedagogically meaningful use, though it does not target specific pedagogical \emph{can-do} statements or grammatical constructs. A more recent example is POLKE (Pedagogically Oriented Language Knowledge Extractor), a rule-based system developed by the KIBi team at the University of Tübingen~\cite{sagirov2025polke}, which we use as one of our rule-based baselines (see Sections \ref{sec:models_feedback} and \ref{sec:models_proficiency}). POLKE automatically annotates 659 grammatical constructs featured in the EGP and is accessible via an API.\footnote{\url{polke.kibi.group/index.html}} In their evaluation of the system, \cite{verratti2025nlp} attempt to assess the validity of the levels associated to the EGP \emph{can-do} statements by applying POLKE to a small subset of the EFCAMDAT corpus~\cite{geertzen2013automatic}, concluding that many of the CEFR levels assigned to constructs in the EGP may be inaccurate. However, several aspects in the study need to be carefully considered. First, POLKE only covers 659 out of the 1,211 \emph{can-do} statements (i.e., only the ones that do not leverage any pragmatic or semantic information), and in practice, only around 550 appear to be triggered in the dataset used by \cite{verratti2025nlp}. Second, POLKE was originally validated using L1 English data~\cite{sagirov2025polke}, i.e., the COCA corpus~\cite{davies2009coca}, with the authors assuming that PoS taggers are sufficiently robust for learner language -- an assumption that is not consistently supported in the literature~\cite{geertzen2013automatic, berzak-etal-2016-universal, chau2023comparison}. In relation to this aspect, another element to take into account is that POLKE may identify the presence of a grammatical construct, but not its correct or incorrect usage. Finally, in \cite{verratti2025nlp}, 
constructs are automatically assigned to the level at which they are most frequently observed in the sample. This 
does not 
fully take into account the cumulative nature of language acquisition, where higher-level learners may demonstrate lower-level constructs, and risks conflating frequency with developmental onset. 

Prior to this study, \cite{kim2021generalizability} examined 35 grammatical constructs associated with A2, B1, and B2 levels, using an L2 English corpus of L1 Korean learners. Their results indicated that construct use generally increased with proficiency, although the overall frequency of these constructs was low. As with \cite{verratti2025nlp}, they also observed some discrepancies between the EGP-assigned levels and actual learner performance. However, they emphasised the influence of task type and L1 background on these outcomes.

The advent of foundation models brought innovative approaches to grammatical constructs detection such as \cite{weissweiler-etal-2022-better}, which investigated the ability of pretrained language models, such as BERT~\cite{devlin2018}, RoBERTa~\cite{liu2019robertarobustlyoptimizedbert}, and DeBERTa~\cite{he2021deberta}, to classify and interpret the English comparative correlative. Using logistic regression on BERT sentence representations, they successfully detected the presence of comparative correlative structures. \cite{okano2023generating} classified a set of constructions similar to those in the EGP using data from SCoRE~\cite{chujo2015corpus} and employing BERT models to detect three constructions: present perfect, relative clauses, and subjunctive. 
These studies demonstrate the potential of transformer-based models in this domain, even if limited to only a few grammatical constructs.

Building on this line of work, \cite{glandorf-meurers-2024-towards} proposed a comprehensive pipeline leveraging LLMs for pedagogically-oriented grammar detection and generation, using the EGP as a reference. They employed LLMs to generate over 940,000 examples for all EGP constructions and used these to train multi-task BERT-based binary classifiers, one per construct, to detect the presence of specific grammatical constructs in sentences. Their approach achieves high performance, although the authors note certain limitations. In particular, the synthetic examples used to train the classifiers tend to be overly homogeneous in structure and lack typical learner errors, raising concerns about generalisability to real-world learner data. Moreover, no manual validation against authentic corpora was performed.

Across most of these studies -- whether focused narrowly on syntactic features or more broadly on grammatical complexity -- a key limitation lies in their heavy, or even exclusive, reliance on PoS taggers and syntactic parsers. As \cite{chau2023comparison} point out, these tools are not always robust, particularly when applied to learner language, which often contains non-canonical structures and errors. Even when highly accurate, such systems are restricted to morpho-syntactic dimensions, leaving out crucial semantic and pragmatic aspects of learner texts. A further limitation is that prior approaches typically stop at detecting the presence of grammatical constructs or extracting morpho-syntactic information, without distinguishing between correct and incorrect uses of grammar.

In this context, we believe that the use of LLMs offers a promising alternative, as they are capable of capturing not only morpho-syntactic~\cite{ide2025make, kennedy-2025-evidence} -- despite some limitations\footnote{\label{fn:cheng}\cite{cheng-amiri-2025-linguistic} showed that LLMs often struggle when used as PoS taggers or dependency parsers; 
making systematic errors in identifying detailed structures such as embedded clauses and verb phrases. To address this limitation, we will describe a strategy to counteract such shortcomings in Section \ref{sec:proficiency_assessment}.} -- but also semantic~\cite{brown2020language} and pragmatic~\cite{sravanthi-etal-2024-pub} dimensions of language, leveraging their pretrained knowledge without necessarily requiring further fine-tuning. For L2 grammatical complexity analysis, this potential becomes even more compelling when LLMs are paired with structured external resources such as the EGP, which provides \emph{can-do} statements, expressed in natural language, targeting highly specific grammatical constructs. To the best of our knowledge, this is the first study to leverage the EGP to detect and classify attempts at grammatical constructs for feedback and to explore the relationship between L2 grammatical complexity and proficiency assessment using an LLM-based approach. Furthermore, unlike previous work, a key novelty of our framework, outlined in the next section, is that it operates on both original and corrected versions of learner sentences, enabling the identification of learners’ attempts at grammatical constructs and their classification as successful or unsuccessful.

\section{Attempts at grammatical constructs}
\label{sec:grammatical_attempt}

\begin{table*}[ht!]
\scriptsize
\centering
\begin{tabular}{cccccc}
\toprule
\multicolumn{2}{c}{\textbf{Text}} & \multicolumn{2}{c}{\textbf{\emph{Can-do} statement}} & \\
\textbf{Original} & \textbf{Correct} & \textbf{In original} & \textbf{In correct} & \textbf{Attempt} & \textbf{Class}\\ \midrule
Music used to be my job.         & Music used to be my job.          & \ding{51}         & \ding{51} & Successful & $\omega_1$        \\ \hline
Music was use to be my job.      & Music used to be my job.          & \ding{55}         & \ding{51} & Unsuccessful   & $\omega_2$     \\ \hline
Music is used to evoke emotion.  & Music is used to evoke emotion.   & \ding{55}         & \ding{55} & No   & \multirow{2}{*}{$\omega_3$}     \\
I used to play music.            & I usually play music.             & \ding{51}         & \ding{55} & Other error         \\ \bottomrule
\end{tabular}
\caption{Grammatical construct attempt labelling framework adopted in our work. In the example, we used EGP 598 (\emph{Can use `used to' to talk about repeated actions or states in the past that are no longer true}).}
\label{tab:framework}
\end{table*}
In our work, the concept of \emph{attempt} revolves around assessing whether a specific grammatical construct, represented by a \emph{can-do} statement, has been applied in a given sentence. Furthermore, if an attempt was made, we wish to assess if it was successful or not. Requiring annotators to directly classify grammatical constructs as successfully attempted, unsuccessfully attempt, or not attempted (i.e., a three-class classification task) would likely have introduced excessive noise due to the subjective nature of such distinctions and the potential for inconsistent interpretations. In order to reduce the cognitive burden of the annotation task, we therefore opted for a binary labelling scheme of dual sets of the original sentence, written by the learner, and a corrected version. Both sentences were labelled as to whether the \emph{can-do} statement has been applied in them (see the supplementary material for the annotation instructions).
Table \ref{tab:framework} shows the four different attempt labelling outcomes that can be obtained from this approach. 
Specifically, 

\begin{itemize}
    \item [1.] if the \emph{can-do} statement is applied in both the original and corrected sentences, it indicates that the associated grammatical construct was successfully realised;
    \item [2.] if the \emph{can-do} statement is absent in the original sentence but appears in the corrected version, it suggests that the learner attempted to use the grammatical construct unsuccessfully;
    \item [3.] if the \emph{can-do} statement is not applied in either the original or the corrected sentence, it means no attempt was made;
    \item [4.] if the \emph{can-do} statement is found in the original sentence but is not retained in the corrected sentence, it indicates that the learner made an error that is unrelated to the use of that particular grammatical construct. In other words, the learner was attempting to mean something different.
\end{itemize}
\noindent
In our experiments, the four outcomes are collapsed into three binary classification tasks (see Table \ref{tab:framework}):

\begin{itemize}
    \item [a.] \textbf{successful} attempt (\(\omega_1\)) vs. rest;
    \item [b.] \textbf{unsuccessful} attempt (\(\omega_2\)) vs. rest;
    \item [c.] \textbf{general} attempt (successful or not) (\(\omega_1\cup\omega_2\)) vs. no attempt (\(\omega_3\)).\footnote{\label{fn:general}We target \(\omega_1 \cup \omega_2\) rather than \(\omega_3\), as it is pedagogically more relevant to identify whether a learner attempted to use a grammatical construct, regardless of success, than to confirm its absence.}
\end{itemize}

\section{Data}

\label{sec:data}

\subsection{English Grammar Profile}
\label{sec:egp}

\begin{figure}[htbp!]
    \centering
    \includegraphics[width=1\linewidth]{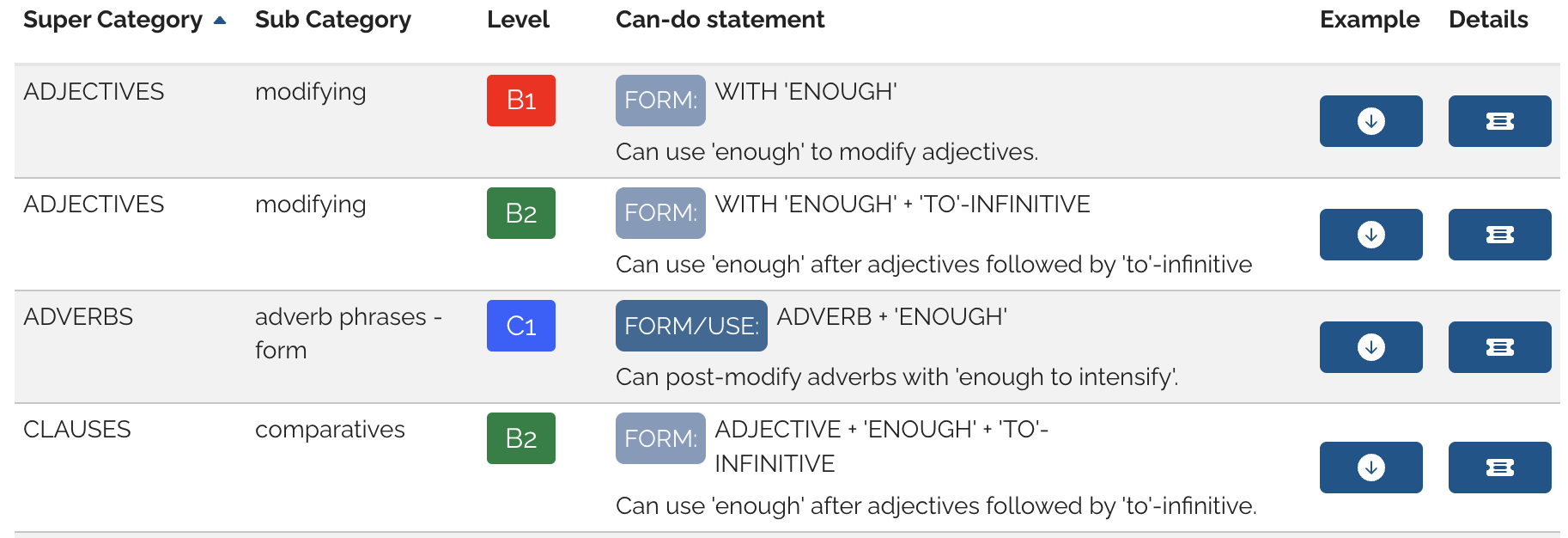}
    \caption{Examples from the EGP website.}
    \label{fig:egp_webpage}
\end{figure}

The English Grammar Profile (EGP) \cite{okeefe2017english} is a publicly available database,\footnote{\url{englishprofile.org/}} which allows users to observe how learners develop competence in grammatical form and meaning, as well as pragmatic appropriateness, as they progress across CEFR proficiency levels. The EGP features a \emph{can-do} statement, its related grammatical categories, the respective CEFR level, and one or more learner examples (see Figure \ref{fig:egp_webpage}). It is important to note that the CEFR levels are not ``exclusive'', but ``cumulative'', e.g., a \emph{can-do} statement at A1 is valid for a C2 speaker (but not vice-versa).

For our experiments on grammatical construct detection and classification, we created a dataset by selecting a subset of an L2 learner essay corpus (see Section \ref{sec:wi2024}) and annotating it using the scheme outlined in Table~\ref{tab:framework}. In this case, we only considered 12 \emph{can-do} statements covering different CEFR levels (A1 to B2) and grammatical categories (adjectives, conjunctions, pronouns, adverbs, nouns, passives, determiners, future, modality, focus, and clauses). In our selection, we prioritised morpho-syntactic variety over proficiency levels, since the initial question that sparked our research focused on whether LLMs can extract information about grammatical constructs expressed in natural language. We also aimed to include constructs that are reasonably frequent in real learner data, which is why we excluded C1 and C2 statements, as these are expected to be rare in the corpus. The selected \emph{can-do} statements can be found in Table \ref{tab:candostat}. For convenience, we adopted a bipartite classification of the selected \emph{can-do} statements based on whether they contain lexical information or not. For example, EGP 37 (\emph{Can use `enough' to modify adjectives}) is a \textbf{lexical} \emph{can-do} statement because it explicitly refers to the word \emph{enough}, whereas EGP 209 (\emph{Can form negative interrogative clauses}) is a \textbf{non-lexical} \emph{can-do} statement. In round brackets, we specify whether a \emph{can-do} statement targets form (FORM), use (USE), or both (FORM/USE).

As mentioned in Section \ref{sec:intro}, we also investigate the predictive power of attempts at grammatical constructs for proficiency assessment. For these experiments (see Section \ref{sec:proficiency_assessment}), we considered all EGP \emph{can-do} statements that included learner examples, totalling 1,211 grammatical constructs. In this case, we used the full development set of the L2 learner essay corpus. Since ground-truth labels for all grammatical construct attempts are not available in this larger dataset, predictions are evaluated indirectly: each detected construct attempt is associated with the CEFR level of the corresponding EGP \emph{can-do} statement and compared against the human-assigned CEFR level of the essay.

\begin{table*}[ht!]
\centering
\scriptsize
\begin{tabular}{
  >{\arraybackslash}p{0.8cm}
  >{\arraybackslash}p{0.5cm}
  >{\arraybackslash}p{8.1cm}
  >{\arraybackslash}p{1.8cm}
  >{\arraybackslash}p{2.8cm}
  >{\arraybackslash}p{0.5cm}
}
\toprule
\textbf{Type} & \textbf{EGP} & \textbf{Text} & \textbf{SuperCategory} & \textbf{SubCategory} & \textbf{CEFR} \\
\midrule

\multirow{11}{*}{\textbf{\rotatebox[origin=c]{90}{Lexical}}} 
 & 37   & Can use \emph{enough} to modify adjectives. (FORM) & adjectives     & modifying                     & B1 \\
 & 266  & Can use \emph{either ... or} to connect two words, phrases or clauses. (FORM) & conjunctions  & coordinating                  & B1 \\
 & 295  & Can use \emph{another} to talk about something different. (USE) & determiners   & articles                      & B1 \\
 & 367  & Can use \emph{would} to talk about the future from a point in the past. (USE) & future        & future in the past            & B1 \\
 & 598  & Can use \emph{used to} to talk about repeated actions or states in the past that are no longer true. (USE) & modality      & used to                       & B1 \\
 & 983  & Can use \emph{everything} as subject, with a singular verb. (FORM) & pronouns      & indefinite -thing, -one, -body & A1 \\
 
\midrule

\multirow{10}{*}{\textbf{\rotatebox[origin=c]{90}{Non-lexical}}}
 & 19   & Can form irregular comparative adjectives. (FORM) & adjectives     & comparatives                  & A2 \\
 & 209  & Can form negative interrogative clauses. (FORM)   & clauses        & interrogatives                & A2 \\
 & 228  & Can use a defining relative clause, without a relative pronoun. (FORM) & clauses        & relative                      & A2 \\
 & 242  & Can use a finite subordinate clause with time conjunctions, before or after a main clause. (FORM/USE) & clauses        & subordinated                  & A2 \\
 & 249  & Can use a finite subordinate clause, before or after a main clause, with conjunctions to introduce conditions. (FORM/USE) & clauses        & subordinated                  & B2 \\
 & 708  & Can use the past simple passive affirmative with a range of pronoun and noun subjects. (FORM) & passives       & passives: form                & B1 \\

\bottomrule
\end{tabular}
\caption{\emph{Can-do} statements considered in the first part of our experiments.}
\label{tab:candostat}
\end{table*}

\subsection{Write \& Improve 2024}
\label{sec:wi2024}

Write \& Improve (W\&I) is an online platform where L2 learners of English can practise their writing skills~\cite{yannakoudakis2018developing}. Users can submit their compositions in response to different prompts, and the W\&I automatic system provides assessment and feedback. Some of these compositions have been annotated with CEFR levels and grammatical error corrections, resulting in a corpus of more than 23,000 essays~\cite{nicholls2024write}.\footnote{\url{https://englishlanguageitutoring.com/datasets/write-and-improve-corpus-2024}}
Here, we used only the final essay versions alongside their corrected counterparts, available both as full texts and sentence splits, enabling parallel comparison between learner-generated and corrected compositions.

For the experiments targeting the detection of attempts at grammatical constructs, we extracted 7,342 sentences (i.e., 3671 sentence pairs) from both the W\&I training and development sets by applying tailored filters specific to each grammatical construct using lexical matching and \texttt{spacy}\footnote{\label{fn:spacy}\url{spacy.io}} on the corrected versions of the sentences. 
The filtering was performed on the corrected sentences, and not on the original learner sentences, to avoid excluding unsuccessful attempts from the analysis (see class \(\omega_{2}\) in Table \ref{tab:framework}). Furthermore, PoS and syntactic taggers perform more reliably on L1-like language, and corrected sentences approximate such norms (see Section~\ref{sec:auto_approaches_feedback} and \cite{berzak-etal-2016-universalfull} specifically).
At the same time, the filter needs to be broad to intentionally include false positives, allowing us to test both the rule-based system's grammatical and syntactic tagger and the LLMs. For example,

\vspace{0.5em}

\noindent
\textit{Lexical}: \noindent \textbf{EGP 37} (\emph{Can use `enough' to modify adjectives}): We filtered for sentences containing the word \emph{enough}, without additional constraints.

\noindent
\textit{Non-lexical}: \noindent \textbf{EGP 228} (\emph{Can use a defining relative clause, without a relative pronoun}): We filtered for sentences containing the \texttt{spaCy} universal dependency \texttt{relcl}, which identifies relative clauses.

\vspace{0.5em}

\noindent
For each construct, if the filter yielded 400 or more sentences, we randomly selected 300; otherwise, we retained all the filtered sentences.
The set of filters applied can be found in the supplementary material.  

After this extraction, we asked a human expert to annotate each sentence with binary labels indicating whether a given \emph{can-do} statement is applied or not.\footnote{The annotation instructions are reported in the supplementary material.} The labels are applied to both the original and corrected sentences, resulting in two sets of binary labels. The prevalence information for each class can be found in Table \ref{tab:prevalence} in Appendix \ref{appendix_other}. Prior to annotating the 3,671 sentence pairs, we conducted a pilot study on 3,600 original sentences (without the respective corrections) drawn from W\&I 2019~\cite{bryant2019bea} (i.e., 300 sentences for each of the 12 selected \emph{can-do} statements), under the assumption that original sentences would be harder to annotate than corrected ones. These sentences were split into two sections, each annotated by two human annotators, while a third annotator annotated all sentences. The average Cohen’s $\kappa$ across all \emph{can-do} statements was 0.92$\pm{0.07}$, indicating very strong agreement. This third annotator then completed the annotations for the dataset used in this study. W\&I 2024 was used for our work instead of W\&I 2019 due to its more recent and diverse dataset, as well as the greater reliability of its scores resulting from updated rater training. In addition to assessing \emph{inter}-annotator agreement in the pilot study, we also evaluated \emph{intra}-annotator agreement, following \cite{abercrombie-etal-2025-consistency}. Specifically, the third annotator re-annotated the sentence pairs approximately nine months after the initial annotation. The Cohen’s $\kappa$ between the first and second annotations across \emph{can-do} statements was 0.97$\pm{0.04}$ for both the original learner sentences and their corrected counterparts. Finally, the annotator reviewed all cases of disagreement and adjudicated a final label. For convenience, we refer to the resulting subset of 3,671 annotated sentence pairs as \textbf{W\&I-EGP}.

For our experiments targeting proficiency prediction, instead, we focused on the development set of W\&I 2024, which comprises 506 essays written in response to 50 different prompts by learners from 17 L1 backgrounds. A breakdown by CEFR level is provided in Table~\ref{tab:wi_2024_stats}. This will be explicitly referred to as \textbf{W\&I development set}.

\begin{table}[ht!]
\scriptsize
\centering

\begin{tabular}{l|r|r|r|r|r|r|r|r|r}
\toprule
\textbf{} & \textbf{A2} & \textbf{A2+} & \textbf{B1} & \textbf{B1+} & \textbf{B2} & \textbf{B2+} & \textbf{C1} & \textbf{C1+} & \textbf{C2} \\
\midrule
\textbf{\#Essays} & 18 & 78 & 87 & 114 & 100 & 53 & 40 & 13 & 3 \\
\textbf{\#Sents.} & 108 & 435 & 566 & 914 & 1030 & 634 & 466 & 169 & 30 \\
\bottomrule
\end{tabular}
\caption{W\&I 2024 statistics.}
\label{tab:wi_2024_stats}
\end{table}

\section{Experimental Setup}
\label{sec:experiment}

\subsection{Grammatical construct detection and classification}
\label{sec:finegrained}

Providing fine-grained feedback (RQ2) requires determining whether learners used a given grammatical construct and whether it was used correctly. To this end, we need to evaluate our systems’ ability to detect and classify such attempts (RQ1). The first part of our experiments focuses on detecting and classifying attempts at grammatical constructs (RQ1), which can then be used to provide fine-grained grammatical feedback (RQ2). Here, we aim to determine whether a given grammatical construct is used by a learner, and whether it is used successfully or unsuccessfully in W\&I-EGP.

\subsubsection{Models for attempt detection and classification}
\label{sec:models_feedback}

We compare two LLMs with POLKE and our own rule-based system.

As mentioned in Sections \ref{sec:intro} and \ref{sec:auto_approaches_feedback}, POLKE~\cite{sagirov2025polke, verratti2025nlp} is a publicly available web-based API built on the Unstructured Information
Management (UIMA) framework~\cite{ferrucci2004uima} that automatically identifies grammatical constructs from the EGP in learner texts. Users submit text via an HTTP request, which is linguistically processed with standard NLP tools (e.g., PoS tagging, parsing, etc.) as well as some custom annotators to simplify downstream processing. The EGP structures are then detected through UIMA Ruta~\cite{kluegl2016uima}, a rule-based framework for information extraction. Each rule is designed to capture a specific construct, based on the corresponding EGP \emph{can-do} statement, its subcategory, and example sentences. In cases where the statement was too vague, additional contextual information (e.g., vocabulary lists such as the Oxford 3000/5000\footnote{\url{https://www.oxfordlearnersdictionaries.com/wordlists/oxford3000-5000}}) was used to refine the rules. Currently, POLKE implements 659 rules covering form-based EGP structures, while use-based constructs are not yet supported (see labels in round brackets in Table \ref{tab:candostat}). The system returns its output as a JSON list, providing the ID of each detected structure along with its position in the input text.

Similarly, for our rule-based system, we used lexical matching, PoS and syntactic tagging to identify our selection of EGP \emph{can-do} statements. For example,

\vspace{0.5em}

\noindent
\textit{Lexical}: \textbf{EGP 37} (\emph{Can use `enough' to modify adjectives}): The rules checks whether the sentence contains a token that satisfies both of the following conditions:

\begin{itemize}
    \item [a.] The token is \emph{enough};

    \item [b.] The token is preceded by a token that is tagged as an adjective (\texttt{ADJ}).
\end{itemize}

\noindent
\textit{Non-lexical}: 
\textbf{EGP 228} (\emph{Can use a defining relative clause, without a relative pronoun}): The rule checks whether the sentence satisfies the following conditions:

\begin{itemize}
\item [a.] The sentence contains a relative clause (i.e., the universal dependency tag \texttt{relcl});
\item [b.] When examining the full clause, no relative pronoun (\emph{that}, \emph{who}, \emph{whom}, or \emph{which}) is present.
\end{itemize}

\noindent
A description of our rule-based system for all 12 constructs is provided in the supplementary material.

For the LLM experiments, we use the open-source Qwen 2.5 32B (4-bit quantised)~\cite{qwen2.5full} and also include GPT-4.1 (\texttt{gpt-4.1})~\cite{openai2023gpt4}\footnote{\url{openai.com/index/gpt-4-1/}} to allow comparison with a larger model. The inclusion of GPT-4.1 is also justifiable from a cost perspective: since this part of the study targets a small set of \emph{can-do} statements, expenses remained manageable, at approximately \$0.002 per sentence pair for a total of roughly \$15.

After querying the LLM about the presence of a given \emph{can-do} statement expressing a given grammatical construct, we extracted the log probabilities of \emph{Yes} and \emph{No} as the first predicted token and applied a softmax over these two values. The resulting probability for \emph{Yes} was used as the model’s confidence that the construct was present.

\subsubsection{Evaluation metrics for grammatical construct detection and classification}

To evaluate the performance of our models in this task, we use Precision, Recall, and $F_{1}$ scores targeting each class, i.e., \emph{general attempt} (\(\omega_1\cup\omega_2\)) which is equivalent to $\neg\omega_3$; see Note \ref{fn:general}), \emph{successful attempt} ($\omega_1$), and \emph{unsuccessful attempt} ($\omega_2$) separately. That is, we treat the problem as three binary (one-vs-rest) classification tasks, one for each class. The calculation process, however, differs from conventional methods as it involves two sets of binary labels (i.e., one for original and one for corrected sentences) instead of one. The reader is also referred to Section \ref{sec:grammatical_attempt} and Table \ref{tab:framework} for the definition of each class.

Specifically, for each \emph{can-do} statement $i$ there are manual labels for each original sentence, $y_{{\tt o}_{i}}\in\{0,1\}$, and its grammatically corrected sentence, $y_{{\tt c}_{i}}\in\{0,1\}$. Using the mapping described in Table~\ref{tab:framework} yields the following manual class label:

\vspace{-1.5em}

\begin{eqnarray}
\label{eq_labels}
y_{i}=\left\{
\begin{array}{l l}
\omega_1, & \text{if }y_{{\tt o}_{i}}=1 {\mbox{ and }} y_{{\tt c}_{i}}=1\\
\omega_2, & \text{if }y_{{\tt o}_{i}}=0 {\mbox{ and }} y_{{\tt c}_{i}}=1\\
\omega_3, & {\mbox{otherwise}}
\end{array}
\right.
\end{eqnarray}


The output from the automated system is an estimated probability that the original sentence used the \emph{can-do} statement $i$, ${\hat p}_{{\tt o}_{i}}$,
and the probability for the corrected sentence, ${\hat p}_{{\tt c}_{i}}$. 
These probabilities are converted to a class label estimate, ${\hat y}_{i}$, using thresholds $\tau_{{\tt o}_{i}}$ and $\tau_{{\tt c}_{i}}$. Thus, at a particular {\it pair} of thresholds:
\begin{eqnarray}
{\hat y}_{i}=\left\{
\begin{array}{l l}
\omega_1, & \text{if }{\hat p}_{{\tt o}_{i}}\geq \tau_{{\tt o}_{i}} {\mbox{ and }} {\hat p}_{{\tt c}_{i}}\geq\tau_{{\tt c}_{i}}\\
\omega_2, & \text{if }{\hat p}_{{\tt o}_{i}}< \tau_{{\tt o}_{i}} {\mbox{ and }} {\hat p}_{{\tt c}_{i}}\geq\tau_{{\tt c}_{i}}\\
\omega_3, & {\mbox{otherwise}}
\end{array}
\right.
\end{eqnarray}
It is now possible to compare these estimated labels, $\hat{y}_{i}$, with the manual labels, $y_{i}$.
As mentioned above, the problem is treated as three binary (one-vs-rest) classification tasks allowing individual feedback for each of the three classes, i.e., successful attempt ($\omega_1$), unsuccessful attempt ($\omega_2$), and general attempt (all instances not in $\omega_3$, i.e., \(\omega_1\cup\omega_2\)).

Compared to standard precision and recall curves for detecting events, there are two thresholds $\tau_{{\tt o}_{i}}$ and $\tau_{{\tt c}_{i}}$. To address this, the upper-bound precision and recall curve is extracted by obtaining the maximum precision (see Figure \ref{fig:egp_249_scatter} as an example, further precision-recall scatterplots can be found in the supplementary material), $\text{MaxPr}(r)$,  for any given recall rate $r$:

\vspace{-1.0em}

\begin{eqnarray}
\text{MaxPr}(r) = \max_{\tau_{{\tt{o}}_{i}}, \tau_{{\tt{c}}_{i}}} \text{Pr}(\tau_{{\tt{o}}_{i}}, \tau_{{\tt{c}}_{i}}) \text{ s.t. } \text{Rc} = r
\end{eqnarray}
F-scores can then be derived from this curve of $\text{MaxP}(r)$ against $r$. Here, $F_{1}$ is used, balancing precision and recall importance.

\begin{figure}[htbp!]
    \centering

    \begin{minipage}{0.72\linewidth}
        \centering
        \includegraphics[width=\linewidth]{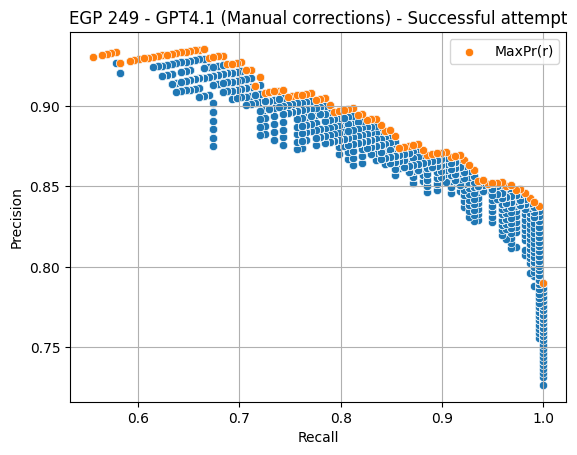}
        \label{fig:egp_249succ}
    \end{minipage}

    \caption{Precision–Recall scatterplot for EGP 249 (\emph{Can use a finite subordinate clause, before or after a main clause, with conjunctions to introduce conditions}). Orange points represent the upper-bound precision and recall obtained through $\text{MaxPr}(r)$.}
    \label{fig:egp_249_scatter}
\end{figure}

As mentioned in Section \ref{sec:intro}, for both the tasks of grammatical construct detection and classification and proficiency assessment, a pipeline using manual corrections is compared to a fully automated pipeline using GECToR~\cite{omelianchuk-etal-2020-gector} (with RoBERTa~\cite{liu2019robertarobustlyoptimizedbert} as the encoder) to generate automatic GEC corrections. For completeness, we also show the results of GECToR by comparing reference and hypothesised edits, and use $F_{0.5}$ score to reflect a weighted precision and recall. As discussed in \cite{ng2014conll}, $F_{0.5}$ is a better choice than $F_{1}$ as higher precision is crucial for user trust in real-world applications.
The reference and hypothesised edits are extracted using the ERRor ANnotation Toolkit (ERRANT)~\cite{bryant2017automatic}, a standard tool used for GEC evaluation.

\subsection{Proficiency assessment}
\label{sec:proficiency_assessment}

At this point, we can establish the validity of our approach for detecting and classifying learners' attempts at grammatical constructs. The second part of our experiments therefore focuses on holistic proficiency assessment, aiming to determine whether attempts at grammatical constructs are reliable predictors of learner proficiency (RQ3) in the W\&I 2024 development set. It is important to note that these experiments are not intended to compete with state-of-the-art automated essay scoring systems, but rather to demonstrate the predictive power of grammatical construct attempts as defined in the EGP. Since we do not have EGP construct annotations for all the sentences in the W\&I 2024 development set, we can only use holistic CEFR labels as reference points. We would normally expect a grammatical construct associated with a given level (e.g., as in Table \ref{tab:candostat}, \emph{Can use ‘enough’ to modify adjectives} at B1) to appear primarily in essays rated at that level or above (e.g., B1, B2, C1, C2), as learners tend to produce constructs they have already mastered, rather than those above their current proficiency level.

\subsubsection{Models for proficiency assessment task}
\label{sec:models_proficiency}

As mentioned in Sections \ref{sec:intro} and \ref{sec:auto_approaches_feedback}, and further detailed in Section \ref{sec:models_feedback}, we used POLKE to detect grammatical constructs, which were then mapped to their respective CEFR proficiency levels for the task of proficiency assessment.

We compare POLKE to a hybrid pipeline that combines a rule-based filtering system with an LLM. For the filtering stage, we implemented 1,211 rules, i.e., one for each EGP \emph{can-do} statement, using lexical matching and \texttt{spaCy}. For example, the rule for EGP 37 (\emph{Can use ``enough'' to modify adjectives}) checks that \emph{enough} is preceded by an adjective according to \texttt{spaCy}’s PoS tags. Where the \emph{can-do} statements themselves specify less restrictive conditions, we deliberately kept our rules broad in order to maximise recall. This was especially necessary for statements requiring semantic interpretation, such as EGP 367 (\emph{Can use ``would'' to talk about the future from a point in the past}), where our rule only checks for the lemma \emph{would} and the presence of a past-tense verb. In short, unlike the rule-based system outlined in Section \ref{sec:models_feedback}, this rule-based filtering system serves only as a lightweight pre-selection mechanism for the LLM, which then provides the actual classification. This two-stage design was essential, since directly querying an LLM for all 1,211 constructs across the full W\&I development set would have been prohibitively costly and time-consuming. To ensure a fair comparison, we will also report results limited to the 659 \emph{can-do} statements covered by POLKE.

For our LLM, we use Qwen 2.5 32B \cite{qwen2.5full}, as in the first part of the experiments described in Section \ref{sec:models_feedback}, which can be efficiently run on two Tesla V100 32GB GPUs. An example of the prompt given to the LLM is included in Appendix \ref{appendix_prompts}. In addition to the target \emph{can-do} statement, the prompt incorporates extra context from the EGP, including the SuperCategory, SubCategory, and corresponding learner examples (see Figure \ref{fig:egp_webpage}). However, as mentioned in Note \ref{fn:cheng}, \cite{cheng-amiri-2025-linguistic} have recently shown that state-of-the-art LLMs struggle to provide fine-grained syntactic and grammatical annotations comparable to those of a standard PoS and universal dependency tagger. We observed similar limitations in our preliminary experiments when prompting the LLM with only this contextual information. To address this, we enriched each target learner sentence with PoS, grammatical, and universal dependency tags (automatically extracted with \texttt{spaCy}), which consistently improved classification performance. Consequently, we incorporated this additional grammatical and syntactic information into the prompt.\footnote{As noted in Section \ref{sec:related}, while the literature cautions that PoS taggers and dependency parsers are not always robust on L2 learner data, here they are not used as a definitive analysis tool but as additional context for the LLM.} The mechanism for extracting the model's confidence that a given construct is present is the same as the one described in Section \ref{sec:models_feedback}.

\subsubsection{Holistic scoring and evaluation metrics for proficiency assessment}
\label{sec:evaluation_proficiency}

Proficiency assessment is evaluated in terms of the ability of the systems to predict the holistic CEFR score for each essay in a test dataset. The general approach taken here to compute the predicted holistic score is as follows.

For each essay \( e \), let \( {\cal S}^{(e)} \) denote the set of its sentences. For each CEFR level, $l$, in \( {\cal L}=\{A1, A2, B1, B2, C1, C2\} \) and associated \emph{can-do} statement \( i \) from the set of all can-do statements of level $l$, ${\cal C}^{(l)}$, and for each sentence \( s \in {\cal S}^{(e)} \), 
an attempt label is predicted 
for the original sentence, $\hat{y}^{(s)}_{{\tt o}_i}\in\{0,1\}$, and its corresponding grammatically corrected sentence, $\hat{y}^{(s)}_{{\tt c}_i}\in\{0,1\}$. Let the predicted presence of a unique attempt of \emph{can}-do statement \( i \) in essay \( e \), for a successful attempt (i.e., \(\omega_1\), following Table~\ref{tab:framework}) be:

\vspace{-0.5em}

\begin{equation*}
\mathds{1}_{e,i} =
\begin{cases}
1 & \text{if } \exists \: s \in {\cal S}^{(e)} : (\hat{y}^{(s)}_{o_{i}} = 1 \wedge \hat{y}^{(s)}_{c_{i}} = 1) \\
0 & \text{otherwise}
\end{cases}
\end{equation*}

\noindent
or for a general attempt (\(\omega_1\cup\omega_2\)):
\begin{equation*}
\mathds{1}_{e,i} =
\begin{cases}
1 & \text{if } \exists \: s \in {\cal S}^{(e)} : \hat{y}^{(s)}_{c_{i}} = 1 \\
0 & \text{otherwise}
\end{cases}
\end{equation*}

Then, to avoid the effect of text length, the total number of unique attempts\footnote{Here, a unique attempt means that if a given grammatical construct is attempted multiple times, we count it only once.} (successful or general) per essay \( e \) and level $l$ are computed and normalised by dividing by the total number of unique attempts per essay \( e \) across all levels. The overall (successful or general) attempt score for essay \( e \) is the weighted sum over CEFR levels \( {\cal L} \), where \( w(l) \) is the numeric weight assigned to each level (e.g., \(w(A1)=1\), \(w(A2)=2\), ..., \(w(C2)=6\)):
\vspace{-1.5em}

\begin{eqnarray}
\hat{{\tt sc}}_{e} = 
\left.\sum_{l \in {\cal L}} w(l)\sum_{i \in {\cal C}^{(l)}}
\mathds{1}_{e,i}
\right/{\sum_{l\in{\cal L}}\sum_{i\in {\cal C}^{(l)}}\mathds{1}_{e,i}}
\end{eqnarray}

The proficiency prediction is evaluated by measuring this attempt-based predicted score against the manual reference holistic CEFR level assigned to the essay in terms of Pearson's correlation coefficient (PCC) and Spearman's rank coefficient (SRC).

This evaluation procedure applies straightforwardly to POLKE. LLMs, however, 
output estimated probabilities of the presence of \emph{can-do} statements in sentences so we need to convert these probabilities first into binary class labels. To do this, we select a threshold from the limited set \(\{0.70, 0.80, 0.90, 0.95, 0.99\}\) for each CEFR level $l$. Although we acknowledge that this is a simplification, this limited set is chosen to reduce computational complexity; otherwise, finding the best combination of thresholds across all six levels would be too time-consuming. For each sentence pair (original and corrected), we assign a positive label if the LLM’s predicted probability exceeds the threshold for that level. To determine the optimal combination of thresholds, we perform a grid search over all possible threshold combinations in the limited set (one threshold per CEFR level $l$) using 5-fold cross-validation. The best combination is selected based on the maximum SRC between the aggregated attempt scores and the holistic CEFR essay scores. Future work will focus on developing improved methods for determining optimal thresholds.

\section{Experimental results}

\subsection{Results for grammatical construct detection and classification}
\label{sec:feedback_results}

In this section, we report the results for grammatical construct detection and classification on W\&I-EGP (RQ1) comparing our rule-based system (RB) to POLKE and two LLMs, i.e., Qwen 2.5 32B and GPT-4.1 (see Section \ref{sec:models_feedback}). Note that POLKE covers only 6 of our 12 selected \emph{can-do} statements.

\begin{figure}[htbp!]
    \centering
    \includegraphics[width=0.75\linewidth]{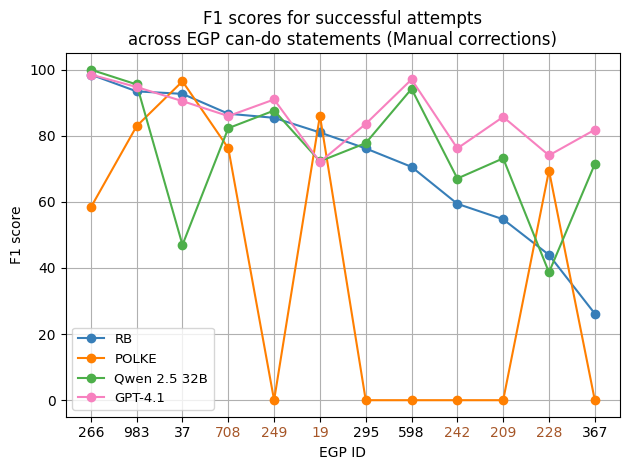}
    \caption{Lineplot of F1 scores for successful attempts across EGP \emph{can-do} statements (Manual corrections).}
    \label{fig:lineplot_f1}
\end{figure}

Figure~\ref{fig:lineplot_f1} shows the $F_{1}$ scores for determining successful attempts across the 12 selected EGP \emph{can-do} statements. 
The $F_{1}$ scores have been ranked according to the RB performance to help guide the reader. Despite the presence of grammatical and syntactic tags in the prompt fed into the LLMs (see the example prompt in Appendix \ref{appendix_prompts}), RB and POLKE -- except for the can-do statements not implemented in the latter -- tend to outperform both the LLMs on constructs that rely strictly on morpho-syntactic cues and require minimal interpretation, e.g., EGP 37 (\emph{Can use ‘enough’ to modify adjectives}), which only involves detecting \emph{enough} following an adjective PoS tag; EGP 19 (\emph{Can form irregular comparative adjectives}), which only requires matching against a fixed list of adjectives (i.e., \emph{further}, \emph{farther}, \emph{better}, \emph{elder}, and \emph{worse}); or -- for RB, specifically -- EGP 708 (\emph{Can use the past simple passive affirmative with a range of pronoun and noun subjects}), which only requires the presence of a passive auxiliary in the past tense (i.e., \emph{was} or \emph{were}). In contrast, as expected, the LLMs perform significantly better on semantically nuanced constructs, such as EGP 295 (\emph{Can use `another' to talk about something different}), EGP 367 (\emph{Can use ‘would’ to talk about the future from a point in the past}), and EGP 598 (\emph{Can use ‘used to’ to describe past habits or states that are no longer true}). Importantly, this should not be taken to mean that LLMs are ineffective at detecting form-based grammatical constructs. For instance, GPT-4.1 outperforms RB on EGP 209 (\emph{Can form negative interrogative clauses}) and surpasses both RB and POLKE on EGP 228 (\emph{Can use a defining relative clause without a relative pronoun}), although Qwen struggles in the latter case. 
The distinction between lexical and non-lexical (highlighted in rust brown in Figure \ref{fig:lineplot_f1}) \emph{can-do} statements does not appear to account for the performance differences observed between the rule-based systems and the LLMs. As expected, given its larger size, GPT-4.1 generally outperforms Qwen 2.5 32B. Complete results for general, successful, and unsuccessful attempts can be found in Table~\ref{tab:individual_feedback_results} in Appendix \ref{appendix_other_res}.

Since POLKE can only be evaluated on the \emph{can-do} statements it implements, Table \ref{tab:manual_polke_only} reports average $F_{1}$ results on its six featured \emph{can-do} statements (i.e., 19, 37, 228, 266, 708, and 983), comparing POLKE, RB, Qwen 2.5 32B, and GPT-4.1 to allow a fair comparison. While, as previously observed in Figure \ref{fig:lineplot_f1}, POLKE outperforms the other models on some individual \emph{can-do} statements (e.g., EGP 19 and 37), on average it surpasses Qwen 2.5 32B for general and successful attempts but underperforms compared to RB and GPT-4.1. For all categories of attempt, GPT-4.1 dominates, followed by RB. However, this evaluation only considers form-based EGP constructs (i.e., those included in POLKE) and is therefore not fully informative.

Therefore, we report the average $F_{1}$ results on all the 12 \emph{can-do} statements in Table~\ref{tab:manual_all_comparison}.\footnote{In the supplementary materials, in addition to $F_{1}$, we also report the complete results in terms of Precision and Recall.} The LLMs outperform RB on average for both lexical and non-lexical \emph{can-do} statements, with GPT-4.1 being the best-performing system. 
The improvement is particularly notable for lexical \emph{can-do} statements, likely due to the presence of clearer lexical cues that aid the LLMs' interpretive capabilities. While the classification of general and successful attempts yields strong results overall, identifying unsuccessful attempts remains noticeably more challenging. In this respect, it is worth noting that the general scarcity of this class across the considered \emph{can-do} statements makes the results quite noisy and not always easy to interpret, especially for threshold-based approaches such as the LLM-based classifiers used here. However, the LLMs still outperform RB in this class. About RB specifically, looking into the reasons for lower performance on detecting unsuccessful attempts, apart from PoS and syntactic tagging errors (see Section \ref{sec:auto_approaches_feedback}), we observed that most misclassifications occur when it predicts $\hat{y}_{o_i} = 1$ and $\hat{y}_{c_i} = 1$ instead of the expected $\hat{y}_{o_i} = 0$ and $\hat{y}_{c_i} = 1$ (see Equation \ref{eq_labels}). This is because RB is designed to trigger a \emph{can-do} statement based on the presence of any element that satisfies its requirements, and therefore cannot handle sentences containing both correct and incorrect occurrences of a grammatical construct, which are labelled as unsuccessful attempts (see annotation instructions in the supplementary material).

A concrete example for EGP 19 (\emph{Can form irregular comparative adjectives}) is the original sentence:

\emph{I love music because it can make us feeling better or badder about our feelings.}

and its corrected version:

\emph{I love music because it can make us feel better or worse about our feelings.}

In this case, RB would trigger the \emph{can-do} statement for both sentences. GPT-4.1, instead, assigns a confidence of 0.5 for the original sentence and 1.0 for the corrected sentence (with thresholds of approximately 0.99 for both original and corrected sentences) reflecting a more accurate distinction between unsuccessful and successful attempts.

\begin{table}[ht!]

\centering
\begin{tabular}{c|ccc}
\toprule
\multirow{2}{*}{\textbf{Model}} &  \multicolumn{3}{c}{\textbf{F$_{1}$}} \\ 
 &  \textbf{General} & \textbf{Successful}  & \textbf{Unsuccessful}\\ 
 \midrule
 \textbf{RB} & 83.52 & 82.70 & 68.45 \\
 \midrule
  \textbf{POLKE} & 78.08 & 78.26 & 52.18 \\
  \midrule
   \textbf{Qwen} & 73.64 & 72.62 & 62.15 \\
   \midrule
\textbf{GPT-4.1} & \textbf{86.36} & \textbf{85.97} & \textbf{68.87} \\

\bottomrule

\end{tabular}
\caption{Average $F_{1}$ scores for \emph{can-do} statements implemented in POLKE for RB, POLKE, Qwen 2.5 32B, and GPT-4.1 (manual corrections).}
\label{tab:manual_polke_only}
\end{table}

\begin{table}[ht!]
\scriptsize
\centering
\begin{tabular}{c|c|ccc}
\toprule
\multirow{2}{*}{\textbf{Statement Type}}& \multirow{2}{*}{\textbf{Model}} &  \multicolumn{3}{c}{\textbf{F$_{1}$}} \\ 
 &  & \textbf{General} & \textbf{Successful}  & \textbf{Unsuccessful}\\ 
 \midrule

\multirow{3}{*}{\textbf{Lexical}}  & RB    & 76.62      & 76.26   & 65.09      \\

 & Qwen   & 81.18    & 80.98     & 73.63   \\ 
  & GPT-4.1   & \textbf{90.46}    & \textbf{91.03}     & \textbf{81.61}    \\ 

\midrule

\multirow{3}{*}{\textbf{Non-lexical}} & RB    & 70.75   & 68.53      & 48.05     \\

 & Qwen    & 72.77     & 70.19    & 54.87     \\
  & GPT-4.1   & \textbf{81.87}    & \textbf{80.83}     & \textbf{65.62}    \\ 

 \midrule
  \midrule

\multirow{3}{*}{\textbf{All}} & RB    & 73.68    & 72.39     & 56.57     \\

 & Qwen   & 76.97    &  75.58   & 64.25   \\
  & GPT-4.1   & \textbf{86.16}    & \textbf{85.93}     & \textbf{73.61}   \\

\bottomrule

\end{tabular}
\caption{Average $F_{1}$ scores for attempts, successful, and unsuccessful attempts for all \emph{can-do} statements for RB, Qwen 2.5 32B, and GPT-4.1 (manual corrections).}
\label{tab:manual_all_comparison}
\end{table}

\begin{table}[ht!]

\centering
\begin{tabular}{c|c|ccc}
\toprule
\multirow{2}{*}{\textbf{Model}}& \multirow{2}{*}{\textbf{Corrections}} &  \multicolumn{3}{c}{\textbf{F$_{1}$}} \\ 
 &  & \textbf{General} & \textbf{Successful}  & \textbf{Unsuccessful}\\ 
 \midrule

\multirow{2}{*}{\textbf{RB}} & Manual    & 73.68    & 72.39       & 56.57      \\

 & GECToR   & 70.38    & 71.81    & 28.80   \\

 \midrule
 \midrule

\multirow{2}{*}{\textbf{Qwen}} & Manual    & 76.97    & 75.58     &  64.25     \\

 & GECToR   & 72.96     & 76.25    & 47.26   \\
\midrule
\multirow{2}{*}{\textbf{GPT-4.1}} & Manual    & \textbf{86.16}     & \textbf{85.93}     & \textbf{73.61}       \\

 & GECToR   & 81.70      & 85.15    & 51.70   \\

\bottomrule

\end{tabular}
\caption{Average $F_{1}$ scores for all \emph{can-do} statements for RB, Qwen 2.5 32B, and GPT-4.1 (manual corrections and GECToR).}
\label{tab:manual_vs_gector}
\end{table}

For practical application, GEC corrections will need to be generated automatically. Table \ref{tab:manual_vs_gector} reports the overall results of replacing manual corrections with GECTOR derived ones.
Table \ref{tab:gector_all_comparison} in Appendix~\ref{appendix_other_res} reports full results analogous to Table~\ref{tab:manual_all_comparison}, this time using GECToR instead of manual corrections.  
Complete results for attempts based on GECToR corrections, analogous to Table \ref{tab:individual_feedback_results}, can be found in the supplementary material.
It can be seen that the pipelines using manual corrections show the best results in almost all cases. However, the fully automated pipeline shows good results for general and successful attempt prediction, to the point that, when using Qwen 2.5 32B, it marginally outperforms the pipeline using manual correction for the task of predicting successful attempts. Conversely, for the task of identifying unsuccessful attempts, we can see a 17 and a 21 point drop in performance for Qwen 2.5 32B and GPT-4.1, respectively. This is likely ascribable to GECToR's tendency to undercorrect. In other words, if the original sentence is labelled as $\hat{y}_{o_i} = 0$ but no edits are performed, its corrected counterpart will be still tagged as $\hat{y}_{c_i} = 0$ (see Equation \ref{eq_labels}), erroneously making it a \emph{no attempt} (i.e., \(\omega_{3}\)). In future work, we plan to explore the use of alternative GEC systems, including those based on LLMs.

To summarise, although the detection of unsuccessful attempts remains more challenging than that of general and successful attempts, our results show that the proposed LLM-based approach -- particularly when using GPT-4.1 -- effectively captures learners’ attempts at grammatical constructs. When combined with the proposed dual binary labelling scheme, the approach reliably distinguishes successful attempts and, to a lesser extent, unsuccessful attempts, thereby addressing RQ1.

Beyond detection and classification, an important open question concerns how this information should be presented to learners in a pedagogically meaningful way (RQ2). One possibility is to integrate grammatical attempt analysis into a holistic automated scoring system. Once a learner’s overall proficiency level has been established, EGP constructs at that level or above that are successfully realised could be used to provide positive feedback, highlighting the learner’s achievements. Constructs above the learner’s level that are attempted but not successfully realised could be used to acknowledge the learner’s effort to use more advanced grammatical structures. Conversely, constructs at or below the learner’s proficiency level that are not successfully realised could be used to identify areas where further revision and practice may be beneficial. Furthermore, we envisage feedback that goes beyond individual grammatical constructs by leveraging the hierarchical structure of the EGP, allowing detected attempts to be linked to SuperCategories and SubCategories (see Figure \ref{fig:egp_webpage}). For instance, if a learner struggles with EGP 19, as in the example discussed above, while also unsuccessfully attempting other comparative-related grammatical constructs, a useful feedback strategy would be to recommend revising the formation of comparative adjectives more generally. Such feedback could be further supported by presenting both original and corrected learner examples drawn from the EGP itself. At the same time, the ability to detect successful attempts allows learners’ progress to be recognised and aligns with our broader goal of emphasising positive, formative feedback.

\subsection{Proficiency assessment results}

In this section, we report the results for proficiency assessment on the W\&I development set, showing PCC and SRC correlations between the predicted grammatical complexity score (as described in Section~\ref{sec:evaluation_proficiency}) and the ground-truth holistic CEFR level (RQ3). We evaluate POLKE and our rule-based filter (RBFilter) acting alone and as a pre-filter to 
Qwen 2.5 32B. Since POLKE covers only 659 of the 1,211 \emph{can-do} statements in the EGP, we compared it against our RBFilter system restricted to the same subset of \emph{can-do} statements. The results are shown in Table~\ref{tab:proficiency_results_polke}.

\begin{table}[ht]
\footnotesize
\centering
\begin{tabular}{l|l|c|c}
\toprule
\textbf{Model} & \textbf{On} & \;\;\textbf{PCC}\;\; & \;\;\textbf{SRC}\;\; \\
\midrule
 \multirow{2}{*}{POLKE} 
        & General           & 0.672  & 0.680  \\
                           & Successful    & 0.711  & 0.718  \\
\midrule
    \multirow{2}{*}{RBFilter (form-based)} 
        & General           & 0.622   & 0.613 \\
                             & Successful  & 0.626   & 0.636  \\

\midrule
    \multirow{2}{*}{POLKE+Qwen} 
        & General attempts           & 0.719   & 0.730  \\
                            & Successful  & 0.754   & 0.769  \\

\midrule
    \multirow{2}{*}{RBFilter (form-based)+Qwen} 
        & General           & 0.726    & 0.733    \\
                             & Successful  & \textbf{0.775}  & \textbf{0.781}  \\

\bottomrule

\end{tabular}
\caption{Proficiency assessment performance on the W\&I development set of grammatical complexity score-based systems based on 659 form-based can-do statements in the EGP and manual GEC corrections.}
\label{tab:proficiency_results_polke}
\end{table}

As explained in Section~\ref{sec:models_proficiency}, our rule-based filtering system is not intended as a standalone model but rather as a lightweight pre-filter designed to maximise recall before passing candidate sentences to the LLM. Consequently, POLKE -- a complete and fully optimised rule-based system -- outperforms our filtering component (RBFilter) when used on its own. However, when the LLM is added, RBFilter+Qwen consistently outperforms POLKE. It also outperforms POLKE as the LLM pre-filter (POLKE+Qwen). This suggests that prioritising recall over precision in the pre-filtering stage allows more potential grammatical construct attempts to reach the LLM (i.e., an average of $\sim$60 per sentence for RBFilter vs. an average of $\sim$25 for POLKE), enabling Qwen to make full use of its interpretative ability and leading to better grammatical complexity assessment.
A consistent finding across all the models 
is that 
applying them only to
successful attempts always yields stronger correlations than using 
general 
attempts, 
supporting the validity of our attempt-based framework.

\begin{table}[ht]

\centering
\begin{tabular}{l|l|c|c}
\toprule
\textbf{GEC} & \textbf{On} & \;\;\textbf{PCC}\;\; & \;\;\textbf{SRC}\;\; \\
\midrule
\multirow{2}{*}{\textbf{{Manual}}} 
     
        & General           & 0.758 & 0.772 \\
                              & Successful  & \textbf{0.791} & \textbf{0.803} \\
    \midrule

\multirow{2}{*}{\textbf{{GECToR}}} 
     
        & General          & 0.765  & 0.782  \\
      & Successful  & 0.783 & 0.799 \\

\bottomrule

\end{tabular}
\caption{Proficiency assessment performance of the RBFilter+Qwen system based on 1,211 can-do statements on the W\&I development set with manual and GECTOR GEC corrections.}
\label{tab:proficiency_results}
\end{table}

Table \ref{tab:proficiency_results} shows the results when considering all the 1,211 \emph{can-do} statements comparing the semi-automated GEC pipeline (i.e., using manual corrections) and the fully automated pipeline (i.e., using corrections generated by GECToR). As expected, the pipeline using manual corrections and successful attempts yields the best overall performance; however, the fully automated counterpart follows closely, trailing by less than one point in both PCC and SRC. All results in Tables \ref{tab:proficiency_results_polke} and \ref{tab:proficiency_results} are statistically significant (\emph{p} $<$0.05). As explained in Section \ref{sec:evaluation_proficiency}, text length, despite being highly correlated with proficiency (PCC=0.820; SRC=0.841), does not impact our results since the number of attempts is normalised. As explained in Section \ref{sec:proficiency_assessment}, these experiments aim to demonstrate the predictive power of the EGP grammatical constructs, not to achieve state-of-the-art essay scoring. In this regard, it is worth noting that even though our approach relies exclusively on grammatical features -- without incorporating other key dimensions typically used in essay scoring, such as vocabulary, coherence and cohesion, or thematic development -- it still achieves strong correlation results.

Table \ref{tab:thresholds} in Appendix \ref{appendix_other_res} reports the optimal threshold configuration for each system. In the same Appendix, for completeness, we report the GEC performance results: Table \ref{tab:gec_performance} shows the results of GECToR's performance for GEC on the W\&I development set compared to the baseline used in \cite{nicholls2024write}, i.e., one-shot Llama 3.1 8B~\cite{llama3}.

\begin{figure}[ht!]
\centering
\begin{minipage}{0.38\textwidth}
\centering
\includegraphics[width=\linewidth]{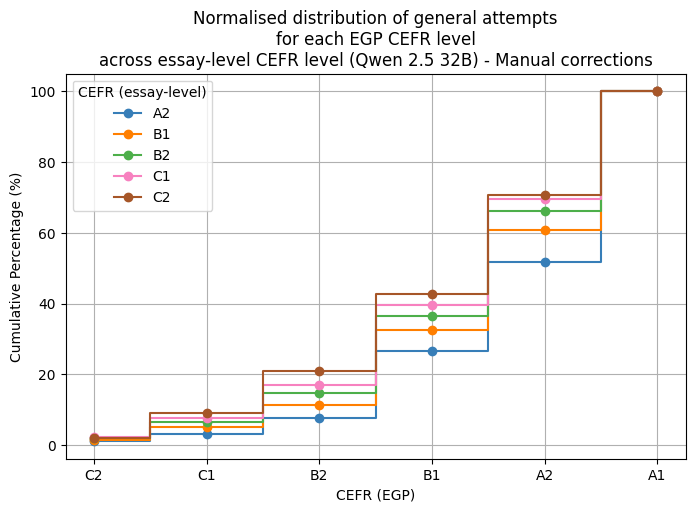}
\end{minipage}
\hfill
\begin{minipage}{0.38\textwidth}
\centering
\includegraphics[width=\linewidth]{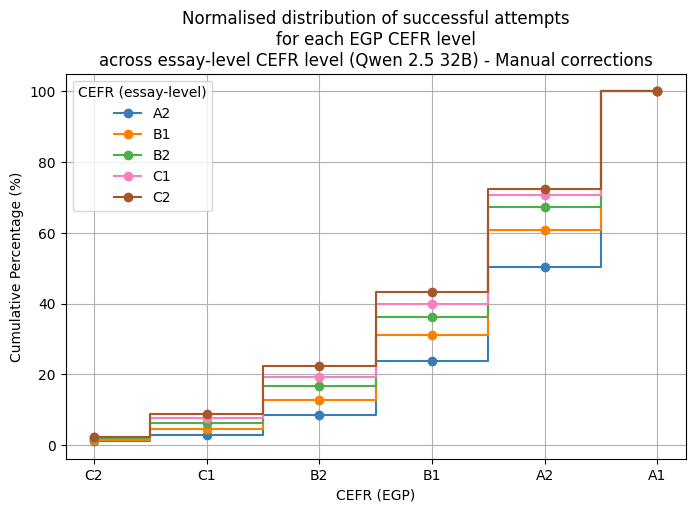}
\end{minipage}
\caption{Cumulative distribution of detected EGP constructs across CEFR levels. For each essay-level CEFR group (A2–C2), the figure shows the cumulative percentage of general (top) and successful (bottom) attempts at or above each CEFR level. RBFilter+Qwen (Manual corrections).}
\label{fig:egp_cumplot}
\end{figure}

Figure \ref{fig:egp_cumplot} shows the cumulative distribution of detected general (top) and successful (bottom) grammatical attempts across CEFR levels, conditioned on essay-level CEFR proficiency. The CEFR levels on the x-axis are reversed to highlight that grammatical complexity is indicative of a given proficiency level or higher, rather than being exclusive to that level (see Section \ref{sec:egp}). Thus, the x-axis represents the grammatical complexity of detected constructions, ordered from C2 (left) to A1 (right). The y-axis indicates the cumulative proportion of detected grammatical attempts that are at least as complex as the corresponding CEFR level. Each curve corresponds to essays of a given human-assigned CEFR level, and each point indicates the proportion of unique grammatical attempts in those essays that meet or exceed the given CEFR level. For example, at B1 on the x-axis, the B2 curve at $\sim$36\% indicates that about 36\% of all detected grammatical attempts in B2 essays are B1-level or higher. As expected, this proportion increases with essay proficiency (e.g., exceeding 40\% for C2 essays) and decreases for lower-proficiency essays (e.g., $\sim$25\% for A2 essays).

\begin{figure}[htbp!]
    \centering
    \includegraphics[width=0.75\linewidth]{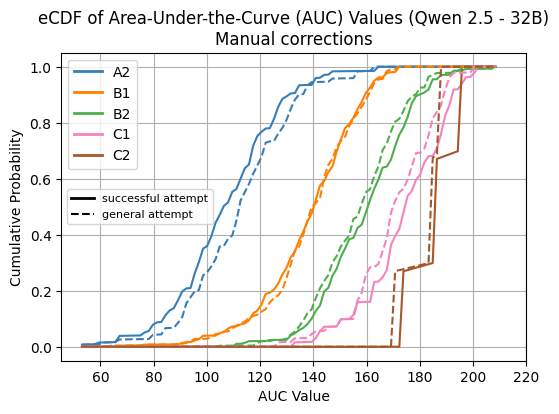}
    \caption{eCDFs of grammar-complexity scores derived from general (dotted) and successful (continuous) attempts at grammatical constructs. Curves are shown for each essay-level CEFR group (A2–C2). RBFilter+Qwen (Manual corrections).}
    \label{fig:egp_ecdf}
\end{figure}

Interestingly, even at higher proficiency levels (e.g., C1 and C2 curves), the use of C1- and C2-level EGP constructs remains relatively infrequent. Nevertheless, the cumulative distributions exhibit clear and systematic distinctions between proficiency bands. To better quantify these differences, Figure \ref{fig:egp_ecdf} presents the empirical cumulative distribution function (eCDF) of the Area Under the Curve (AUC) values derived from Figure \ref{fig:egp_cumplot}. The AUC provides a scalar summary of each essay’s cumulative grammar-complexity profile, with higher values indicating a greater concentration of grammatical mass at higher CEFR levels. The x-axis shows the AUC value, while the y-axis represents the proportion of essays whose AUC is less than or equal to a given value.

As expected, essays at higher CEFR levels exhibit systematically higher grammar-complexity scores, with their AUC distributions shifting rightward as proficiency increases. For instance, when considering successful attempts (continuous lines), at an AUC value of 120, $\sim$78\% of A2 essays (blue curve) have AUC $\leq$ 120, whereas $\sim$2\% of B2 essays fall below this threshold. Clear separations, particularly between A2, B1, and B2, indicate strong differentiation between proficiency levels. The C2 curve (shown in rust brown) appears heavily quantised due to the fact that the W\&I development set contains only three essays at this level.

Furthermore, comparing grammar-complexity distributions based on general (continuous line) versus successful (dotted line) attempts reveals systematic proficiency-dependent differences. At lower proficiency levels (A2), learners frequently attempt constructions beyond their reliable control, resulting in lower complexity scores when only successful uses are considered. At intermediate proficiency (B1), attempted and successful constructions largely coincide, suggesting consolidated grammatical control~\cite{richards2008moving, hawkins2010}. At higher proficiency levels (B2–C2), learners attempt increasingly complex constructions, but not all attempts are fully successful, leading to a modest divergence between the two measures, with complexity scores being slightly higher for successful than general attempts.

In light of the results reported in this section, we show that EGP constructs can be effectively used for proficiency assessment (RQ3). In particular, proficiency predictions are more accurate when such constructs are used successfully by learners than when they are merely attempted. Furthermore, consistent with and corroborating the findings in Section~\ref{sec:feedback_results}, incorporating an LLM into a hybrid pipeline yields substantial improvements over relying on a rule-based system alone.

\section{Conclusions}

In this work, we proposed a novel framework for analysing L2 grammatical complexity based on \emph{can-do} statements from the English Grammar Profile. Central to our approach is the notion of \emph{grammatical attempts}, which allows us to move beyond traditional binary judgments of correctness and to distinguish between general, successful, and unsuccessful uses of grammatical constructs. This distinction enables a more fine-grained characterisation of learners’ grammatical competence, capturing not only what learners can already control, but also where they are actively experimenting with structures beyond their current level.

Our experimental results show that grammatical construct detection and classification can be performed reliably using automatic systems (RQ1). Rule-based systems remain highly effective for form-driven, morpho-syntactic constructions, while LLMs excel at interpreting semantically and pragmatically nuanced uses. However, considering average performance, LLM-based approaches outperform rule-based systems.

Furthermore, we demonstrated how attempt-based grammatical information can support pedagogically meaningful feedback (RQ2). By linking detected attempts to the hierarchical structure of the EGP, feedback can be aggregated beyond individual constructions to broader grammatical categories, enabling learners to identify both persistent difficulties and emerging strengths. Crucially, the ability to detect successful attempts supports a formative feedback paradigm that emphasises learners’ progress and achievements rather than focusing solely on errors.

Our results also show that grammatical complexity scores derived from detected EGP constructs correlate strongly with holistic CEFR proficiency levels (RQ3). Proficiency assessment is consistently more accurate when based on successful attempts rather than on general attempts. Furthermore, a recall-oriented rule-based pre-filter combined with an LLM yields the strongest performance.

We also compared semi-automated pipelines relying on manual grammatical corrections with fully automated pipelines based on GECToR. While manual corrections yield the strongest results overall, the fully automated setting performs competitively for general and successful attempt detection, indicating that the proposed framework is viable in realistic deployment scenarios. At the same time, the reduced performance for unsuccessful attempts highlights the importance of further research into more accurate grammatical error correction modules.

Overall, this work shows that EGP-based grammatical constructs, when analysed through the lens of learner attempts, provide a powerful and interpretable basis for both fine-grained feedback and proficiency assessment. Future work will explore extending the framework to spoken learner data and investigating alternative GEC systems, including LLM-based GEC, to further improve robustness across modalities.

\section*{Limitations}
\label{sec:limitations}

First, we acknowledge that our data selection process to build W\&I-EGP may present a limitation. This targeted extraction was necessary because manually annotating the entire original W\&I 2024 corpus would have been prohibitively costly. Additionally, randomly sampling a fixed number of sentences across all \emph{can-do} statements would likely have resulted in the majority being labelled as \emph{no attempt} (i.e., \(\omega_{3}\)). However, this selection method introduces a substantial bias toward the \emph{general attempt} class (i.e., \(\omega_1\cup\omega_2\)). Consequently, the classification task of distinguishing \emph{general attempt} from \emph{no attempt} becomes more challenging than it would be under a distribution representative of real-world data. Likewise, this process likely favours sentences annotated as \emph{successful attempt} (i.e., \(\omega_{1}\)) over those marked as \emph{unsuccessful attempt} (i.e., \(\omega_{2}\)).

With respect to the experiments on proficiency assessment (see Section~\ref{sec:proficiency_assessment}), we note an important simplification: in order to reduce computational complexity, we selected a single threshold per CEFR level from a limited set of candidate values. Ideally, we would prefer to select thresholds at a more granular level, i.e., for each \emph{can-do} statement individually and, more precisely, separate thresholds for original learner sentences and their corrected counterparts in the case of successful attempts. We acknowledge that this approximation entails a loss of potentially valuable information, and we plan to explore more fine-grained thresholding strategies in future work. It is also important to clarify that this set of experiments is not intended to achieve state-of-the-art essay scoring performance, but rather to demonstrate that grammatical constructs function as criterial features of L2 proficiency.

Another limitation is the limited scope of \emph{can-do} statements used in the section of our experiments related to grammatical construct detection and classification. This part of the study includes only 12 out of the over 1,000 available in the English Grammar Profile, covering proficiency levels from A1 to B2. While this still represents a broad range, it does not encompass the full spectrum of language proficiency. In constructing this subset, our primary objective was to investigate methods for automatically extracting information about grammatical constructs; accordingly, we prioritised diversity in grammatical categories (see Section~\ref{sec:data}) rather than coverage across proficiency levels. We plan to expand the coverage by incorporating additional \emph{can-do} statements, including those targeting higher proficiency levels. To partially compensate for this limitation, however, the second part of our study, albeit in relation to proficiency, does consider and indirectly evaluate all available \emph{can-do} statements, thus offering a more comprehensive overview.

\section*{Acknowledgments}
This paper reports on research supported by Cambridge University Press \& Assessment, a department of The Chancellor, Masters, and Scholars of the University of Cambridge.

\appendices

\section{Class Prevalence Statistics}
\label{appendix_other}

Table \ref{tab:prevalence} reports the prevalence information of each class across all the EGP \emph{can-do} statements considered in our experiments on grammatical construct detection and classification.

\begin{table}[ht!]
\scriptsize

    \centering
    \begin{tabular}{c|c|c||c|c|cc}
        \hline
        & \textbf{EGP} & \textbf{\#Sents.} & \textbf{Succ.} & \textbf{Unsucc.} & \multicolumn{2}{c}{\textbf{No attempt}} \\
        \multirow{12}{*}{\textbf{\rotatebox[origin=c]{90}{Lex.}}} & \textbf{} & \textbf{} & \textbf{} & \textbf{} & \textbf{No attempt} & \textbf{Other error} \\ \hline
         & 37  & 300  & 14.3  & 2.3  & 83.3  & 0.0   \\ \cmidrule(lr){2-7}
         & 266  & 96   & 35.4 & 9.4 & 55.2  & 0.0   \\ \cmidrule(lr){2-7}
         & 295 & 300   & 54.3  &  6.0    &  39.7      &  0.0     \\ \cmidrule(lr){2-7}
         & 367  & 363  & 6.9  & 2.2  & 90.9  & 0.0   \\ \cmidrule(lr){2-7}
         & 598  & 396  & 21.5  & 1.5  & 76.3  & 0.7   \\ \cmidrule(lr){2-7}
         & 983  & 319  & 21.0 & 5.9  & 73.0  & 0.0  \\ \cmidrule(lr){2-7}
         \hline
         \multirow{10}{*}{\textbf{\rotatebox[origin=c]{90}{Non lex.}}} 
         & 19  & 300  & 35.0  & 3.0  & 61.7  & 0.3   \\ \cmidrule(lr){2-7}
         & 209  & 300  & 18.7  & 9.0  & 72.3  & 0.0  \\ \cmidrule(lr){2-7}
         & 228  & 300  & 14.0  & 1.0  & 85.0  & 0.0   \\ \cmidrule(lr){2-7}
         & 242  & 300  & 20.0   & 1.7   & 78.3   & 0.0    \\ \cmidrule(lr){2-7}
         & 249  & 300   & 72.7   & 3.3    & 24.0   & 0.0     \\ \cmidrule(lr){2-7}
         & 708  & 397  & 53.6  & 14.1  & 31.7  & 0.5   \\ \cmidrule(lr){1-7}
    \end{tabular}
    \caption{Classes breakdown - Prevalence (\%).}
    \label{tab:prevalence}
\end{table}

\section{Prompt}
\label{appendix_prompts}

This is an example prompt used for our experiments on both grammatical construct detection and classification and proficiency assessment. The parts of the prompt shown in bold denote the prompt skeleton, which is kept constant across different input sentences and \emph{can-do} statements.
\vspace{1em}

{\small
\noindent
    \texttt{\textbf{Read this sentence written by an L2 learner of English and its respective PoS, grammatical, and universal dependency tags associated to each token:}}\\
\texttt{'Without a high English level, it would be impossible to get a job or to continue my further studies.'}\\
\texttt{[('Without', 'ADP', 'IN', 'prep'), ('a', 'DET', 'DT', 'det'), ('high', 'ADJ', 'JJ', 'amod'), ('English', 'ADJ', 'JJ', 'amod'), ('level', 'NOUN', 'NN', 'pobj'), (',', 'PUNCT', ',', 'punct'), ('it', 'PRON', 'PRP', 'nsubj'), ('would', 'AUX', 'MD', 'aux'), ('be', 'AUX', 'VB', 'ROOT'), ('impossible', 'ADJ', 'JJ', 'acomp'), ('to', 'PART', 'TO', 'aux'), ('get', 'VERB', 'VB', 'xcomp'), ('a', 'DET', 'DT', 'det'), ('job', 'NOUN', 'NN', 'dobj'), ('or', 'CCONJ', 'CC', 'cc'), ('to', 'PART', 'TO', 'aux'), ('continue', 'VERB', 'VB', 'conj'), ('my', 'PRON', 'PRP\$', 'poss'), ('further', 'ADJ', 'JJ', 'amod'), ('studies', 'NOUN', 'NNS', 'dobj'), ('.', 'PUNCT', '.', 'punct')]}\\
\texttt{\textbf{Does the following can-do statement apply to this sentence? Just answer Yes or No without adding any comments, notes, or explanations. The SuperCategory, SuperCategory, and Guideword entries will help you contextualise the can-do statement better. Furthermore, you will see one or more examples written by other L2 learners for which the can-do statement applies.}}\\
\texttt{\textbf{Can-do statement:} Can form irregular comparative adjectives.}\\
\texttt{\textbf{SuperCategory:} ADJECTIVES}\\
\texttt{\textbf{SubCategory:} comparatives}\\
\texttt{\textbf{Guideword:} FORM: IRREGULAR}\\
\texttt{\textbf{Example(s):}}\\
\texttt{What colour do you think is better?}\\
\texttt{For further information, contact Joey Hung.}\\
\texttt{\textbf{Your answer:}}

}

\section{Additional results}
\label{appendix_other_res}

Table \ref{tab:thresholds} reports the optimal threshold configuration across CEFR levels for the hybrid systems (i.e., RBFilter + LLM) used for proficiency assessment.

\begin{table}[ht!]
\scriptsize
\centering
\begin{tabular}{c|c|c|c|c|c|c|c}
\hline
\textbf{GEC} & \textbf{On} & \textbf{A1} & \textbf{A2} & \textbf{B1} & \textbf{B2} & \textbf{C1} & \textbf{C2} \\

\hline
\multirow{2}{*}{\textbf{Manual}} & General & 0.50 & 0.99 & 0.99 & 0.99 & 0.95 & 0.95 \\
\cmidrule(lr){2-8}
 & Successful & 0.50 & 0.90 & 0.99 & 0.50 & 0.90 & 0.80 \\
\midrule
\multirow{2}{*}{\textbf{GECToR}} & General & 0.50 & 0.95 & 0.99 & 0.50 & 0.60 & 0.90 \\
\cmidrule(lr){2-8}
 & Successful & 0.50 & 0.50 & 0.99 & 0.50 & 0.99 & 0.90 \\
 \bottomrule
\end{tabular}
\caption{Optimal thresholds configuration for the hybrid proficiency assessment systems.}
\label{tab:thresholds}
\end{table}

GECToR's performance on the W\&I development set compared to the baseline used in \cite{nicholls2024write} is shown in Table \ref{tab:gec_performance}.

\begin{table}[ht!]
\scriptsize
\centering
\begin{tabular}{lccc}
\toprule
\textbf{Model} & \textbf{Precision} & \textbf{Recall} & \bm{$F_{0.5}$} \\
\midrule
\cite{nicholls2024write} & 39.3  & 42.2  & 39.8  \\
\hline
GECToR & 49.0  & 40.3  & \textbf{46.9}  \\

\bottomrule
\end{tabular}
\caption{GEC performance on W\&I development set.}
\label{tab:gec_performance}
\end{table}

Table \ref{tab:gector_all_comparison} reports the average ${F_{1}}$ performance considering lexical, non-lexical, and all \emph{can-do} statements using RB, Qwen 2.5 32B, and GPT-4.1 based on GECToR corrections.
Table \ref{tab:individual_feedback_results} reports the $F_{1}$ scores for each of the 12 selected \emph{can-do} statements using manual corrections. The corresponding table using GECToR corrections is in the supplementary material.




\begin{table}[ht!]
\scriptsize
\centering
\begin{tabular}{c|c|ccc}
\toprule
\multirow{2}{*}{\textbf{Type}}& \multirow{2}{*}{\textbf{Model}} &  \multicolumn{3}{c}{\textbf{F$_{1}$}} \\ 
 &  & \textbf{General} & \textbf{Successful}  & \textbf{Unsuccessful}\\ 
 \midrule

\multirow{3}{*}{\textbf{Lexical}}  & RB    & 71.51       & 75.32  & 29.60      \\

 & Qwen   & 74.56    & 80.91     & 56.41   \\ 
  & GPT-4.1   & \textbf{84.38}    & \textbf{90.32}      & \textbf{56.47}     \\ 

\midrule

\multirow{3}{*}{\textbf{Non-lexical}} & RB    & 69.25    & 68.30      & 28.01     \\

 & Qwen    & 71.36     & 71.60    &  38.11    \\
  & GPT-4.1   & \textbf{79.02}     & \textbf{79.98}      & \textbf{46.94}    \\ 

 \midrule
  \midrule

\multirow{3}{*}{\textbf{All}} & RB    & 70.38    & 71.81      & 28.80    \\

 & Qwen   &  72.96    & 76.25    & 47.26   \\
  & GPT-4.1   & \textbf{81.70}    & \textbf{85.15}      & \textbf{51.70}    \\

\bottomrule

\end{tabular}
\caption{Average $F_{1}$ scores for all \emph{can-do} statements for RB, Qwen 2.5 32B, and GPT-4.1 (GECToR corrections).}
\label{tab:gector_all_comparison}
\end{table}


\begin{table}[ht!]
\scriptsize
\centering

\begin{tabular}{c|c|c|c|c|c}
\hline
& \textbf{EGP} & \textbf{Model} & \textbf{General} & \textbf{Successful} & \textbf{Unsuccessful} \\
\hline
\multirow{24}{*}{\rotatebox{90}{\textbf{Lexical}}} & \multirow{4}{*}{37}  & RB & 92.63 & 92.68   & \textbf{92.31}   \\ \cmidrule(lr){3-6}
 & & POLKE & \textbf{95.92}  & \textbf{96.47} & \textbf{92.31} \\ \cmidrule(lr){3-6}
                    &  & Qwen & 48.13  & 46.98  & 44.44     \\ 
                     \cmidrule(lr){3-6}
& & GPT-4.1 & 90.72 & 90.48 & 66.67 \\ \cmidrule(lr){2-6}

& \multirow{4}{*}{266}  & RB & \textbf{100.00}  & 98.51     & 94.74    \\ \cmidrule(lr){3-6}
& & POLKE & 54.24  & 58.33 & 18.18 \\ \cmidrule(lr){3-6}
                    &  & Qwen & \textbf{100.00}      & \textbf{100.00}     & \textbf{100.00}  \\  
                    \cmidrule(lr){3-6}
                    & & GPT-4.1 & \textbf{100.00} & 98.51 & 94.74 \\ \cmidrule(lr){2-6}

& \multirow{4}{*}{295}  & RB & 75.26  & 76.17     & 56.60    \\ \cmidrule(lr){3-6}
& & POLKE & -  & - & - \\ \cmidrule(lr){3-6}
                    &  & Qwen & 76.27      & 77.78     & 63.16  \\  
                    \cmidrule(lr){3-6}
                    & & GPT-4.1 & \textbf{81.95} & \textbf{83.59} & \textbf{65.00} \\ \cmidrule(lr){2-6}

& \multirow{4}{*}{367}  & RB & 26.05   & 26.19   & 25.53   \\ \cmidrule(lr){3-6}
& & POLKE & -  & - & - \\ \cmidrule(lr){3-6}
                    &  & Qwen & 70.59    & 71.43   & 66.67      \\ 
                    \cmidrule(lr){3-6}
                    & & GPT-4.1 & \textbf{78.26} & \textbf{81.82} & \textbf{75.00} \\ \cmidrule(lr){2-6}

& \multirow{4}{*}{598}  & RB & 70.31   & 70.59   & 44.44  \\ \cmidrule(lr){3-6}
& & POLKE & -  & - & - \\ \cmidrule(lr){3-6}
                    &  & Qwen & 95.56    & 94.19     & 83.33  \\ 
                   \cmidrule(lr){3-6}
                   & & GPT-4.1 & \textbf{97.83} & \textbf{97.04} & \textbf{100.00} \\ \cmidrule(lr){2-6}

& \multirow{4}{*}{983}  & RB & 95.45     &  93.43    & 76.92    \\ \cmidrule(lr){3-6}
& & POLKE & 86.08  & 82.93 & 74.29 \\ \cmidrule(lr){3-6}
                    &  & Qwen & \textbf{96.55}     & \textbf{95.52}     & 84.21    \\ \cmidrule(lr){3-6}
                    & & GPT-4.1 & 94.05 & 94.74 & \textbf{88.23} \\ \hline

\multirow{24}{*}{\rotatebox{90}{\textbf{Non-lexical}}} & \multirow{4}{*}{19}  & RB & 82.19  & 81.00     & \textbf{73.68}    \\ \cmidrule(lr){3-6}
& & POLKE & \textbf{86.31}   & \textbf{86.10} & 55.56 \\ \cmidrule(lr){3-6}
                    &  & Qwen & 72.73      & 72.28    & 57.14  \\  \cmidrule(lr){3-6}
                    & & GPT-4.1 & 72.34 & 72.07 & 60.87 \\ \cmidrule(lr){2-6}

& \multirow{4}{*}{209}  & RB & 64.34     & 54.73     & 31.58   \\ \cmidrule(lr){3-6}
& & POLKE & -  & - & - \\ \cmidrule(lr){3-6}
                    &  & Qwen & 82.76      & 73.17    & 50.00       
                       \\ \cmidrule(lr){3-6}
                       & & GPT-4.1 & \textbf{90.32} & \textbf{85.71} & \textbf{65.12} \\ \cmidrule(lr){2-6}

& \multirow{4}{*}{228}  & RB & 41.79   & 43.96   & 0.00   \\ \cmidrule(lr){3-6}
& & POLKE & 66.15  & 69.49 & 0.00 \\ \cmidrule(lr){3-6}
                    &  & Qwen & 39.15    & 38.67   & 25.00      \\ \cmidrule(lr){3-6}
                    & & GPT-4.1 & \textbf{71.91} & \textbf{74.07} & \textbf{33.33} \\ \cmidrule(lr){2-6}

& \multirow{4}{*}{242}  & RB & 60.75   & 59.41   & 50.00   \\ \cmidrule(lr){3-6}
& & POLKE & -  & - &  -\\ \cmidrule(lr){3-6}
                    &  & Qwen & 68.21     & 67.07    & 57.14   \\ \cmidrule(lr){3-6}
                
                    & & GPT-4.1 & \textbf{76.43} & \textbf{76.19} & \textbf{75.00} \\ \cmidrule(lr){2-6}

& \multirow{4}{*}{249}  & RB & 86.36   & 85.43    & 60.00    \\ \cmidrule(lr){3-6}
& & POLKE & -  & - & - \\ \cmidrule(lr){3-6}
                    &  & Qwen & 88.50     & 87.63     & 77.78  \\ \cmidrule(lr){3-6}
                    & & GPT-4.1 & \textbf{91.09} & \textbf{90.98} & \textbf{90.00} \\ \cmidrule(lr){2-6}

& \multirow{4}{*}{708}  & RB & 89.08     & \textbf{86.67}     & \textbf{73.04}    \\ \cmidrule(lr){3-6}
& & POLKE & 79.77  & 76.24 & 72.73 \\ \cmidrule(lr){3-6}
                    &  & Qwen & 85.28     & 82.30     & 62.14    \\ \cmidrule(lr){3-6}
                    & & GPT-4.1 & \textbf{89.16} & 85.96 & 69.39 \\ \hline

\end{tabular}

\captionof{table}{Results for RB, POLKE, Qwen 2.5 32B, and GPT-4.1 for grammatical construct detection and classification in terms of $F_{1}$ score (manual corrections).}
\label{tab:individual_feedback_results}
\end{table}




\clearpage

\bibliographystyle{IEEEtran}
\bibliography{refs}

@inproceedings{okano2023generating,
  title={Generating dialog responses with specified grammatical items for second language learning},
  author={Okano, Yuki and Funakoshi, Kotaro and Nagata, Ryo and Okumura, Manabu},
  booktitle={Proc. of the 18th Workshop on Innovative Use of NLP for Building Educational Applications (BEA 2023)},
  pages={184--194},
  year={2023}, doi={10.18653/v1/2023.bea-1.16}
}

@misc{nicholls2024write,
  author = {Diane Nicholls and Andrew Caines and Paula Buttery},
  year = {2024},
  title = {The {W}rite \& {I}mprove {C}orpus 2024: Error-annotated and {CEFR}-labelled essays by learners of {E}nglish},
  publisher = {Cambridge University Press & Assessment},
  url = {https://doi.org/10.17863/CAM.112997},
doi={https://doi.org/10.17863/CAM.112997}
}

@book{cefr2001,
author= {{Council of Europe}},
title={Common European Framework of Reference for Languages: Learning, Teaching, Assessment},
address={Cambridge},
publisher={Cambridge University Press},
year=2001,
url={https://rm.coe.int/1680459f97}
}

@inproceedings{devlin2018,
    title = "{BERT}: Pre-training of Deep Bidirectional Transformers for Language Understanding",
    author = "Devlin, Jacob  and
      Chang, Ming-Wei  and
      Lee, Kenton  and
      Toutanova, Kristina",
    booktitle = "Proc. of the 2019 Conference of the North {A}merican Chapter of the Association for Computational Linguistics: Human Language Technologies, Volume 1 (Long and Short Papers)",
    month = jun,
    year = "2019",
    url = "https://aclanthology.org/N19-1423",
    doi = "10.18653/v1/N19-1423",
    pages = "4171--4186",
}

@article{brown2020language,
  title={Language models are few-shot learners},
  author={Brown, Tom and Mann, Benjamin and Ryder, Nick and others},
  journal={Advances in Neural Information Processing Systems},
  volume={33},
  pages={1877--1901},
  year={2020}
}

@article{kerr2020feedback,
title={Giving feedback to language learners.},
author={Philip Kerr},
journal={Part of the Cambridge papers in ELT series},
address={Cambridge},
publisher={Cambridge University Press},
year={2020},
pages={1-28}
}

@misc{liu2019robertarobustlyoptimizedbert,
      title={{RoBERTa}: A Robustly Optimized {BERT} Pretraining Approach}, 
      author={Yinhan Liu and Myle Ott and Naman Goyal and others},
      year={2019},
      eprint={1907.11692},
      archivePrefix={arXiv},
      primaryClass={cs.CL},
      url={https://arxiv.org/abs/1907.11692}, 
}

@misc{llama3,
      title={The {Llama} 3 Herd of Models}, 
      author={{Llama Team}},
      year={2024},
      url={https://ai.meta.com/research/publications/the-llama-3-herd-of-models/}, 
}

@misc{openai2023gpt4,
      title={{GPT-4 Technical Report}}, 
      author={OpenAI},
      year={2023},
      eprint={2303.08774},
      archivePrefix={arXiv},
      primaryClass={cs.CL},
      doi={10.48550/arXiv.2303.08774}
}

@inproceedings{yancey-etal-2023-rating,
    title = "Rating Short {L}2 Essays on the {CEFR} Scale with {GPT}-4",
    author = "Yancey, Kevin P.  and
      Laflair, Geoffrey  and
      Verardi, Anthony  and
      Burstein, Jill",
    booktitle = "Proc. of the 18th Workshop on Innovative Use of NLP for Building Educational Applications (BEA 2023)",
    month = jul,
    year = "2023",
    url = "https://aclanthology.org/2023.bea-1.49",
    doi = "10.18653/v1/2023.bea-1.49",
    pages = "576--584",
}

@article{hamplyons1995rating,
 ISSN = {00398322},
 URL = {http://www.jstor.org/stable/3588173},
 author = {Liz Hamp-Lyons},
 journal = {TESOL Quarterly},
 number = {4},
 pages = {759--762},
 publisher = {[Wiley, Teachers of English to Speakers of Other Languages, Inc. (TESOL)]},
 title = {Rating Nonnative Writing: The Trouble with Holistic Scoring},
 volume = {29},
 year = {1995},
doi={https://doi.org/10.2307/3588173}
}

@book{weigle2002assessing, place={Cambridge}, series={Cambridge Language Assessment}, title={Assessing Writing}, publisher={Cambridge University Press}, author={Weigle, Sara Cushing}, year={2002}, collection={Cambridge Language Assessment}, address={Cambridge}, doi={https://doi.org/10.1017/CBO9780511732997}}

@article{hawkins2010, title={Criterial Features in Learner Corpora: Theory and Illustrations}, volume={1}, DOI={10.1017/S2041536210000103}, journal={English Profile Journal}, publisher={Cambridge University Press}, author={Hawkins, John A. and Buttery, Paula}, year={2010}, pages={1--23}}

@article{ortega2003syntactic,
  title={{Syntactic complexity measures and their relationship to L2 proficiency: A research synthesis of college-level L2 writing}},
  author={Ortega, Lourdes},
  journal={Applied Linguistics},
  volume={24},
  number={4},
  pages={492--518},
  year={2003},
  publisher={Oxford University Press},
doi={https://doi.org/10.1093/applin/24.4.492}
}

@article{iwashita2006syntactic,
  title={{Syntactic complexity measures and their relation to oral proficiency in Japanese as a foreign language}},
  author={Iwashita, Noriko},
  journal={Language Assessment Quarterly: An International Journal},
  volume={3},
  number={2},
  pages={151--169},
  year={2006},
  publisher={Taylor \& Francis},
doi={https://doi.org/10.1207/s15434311laq0302_4}
}

@inproceedings{chen-meurers-2016-ctap,
    title = "{CTAP}: A Web-Based Tool Supporting Automatic Complexity Analysis",
    author = "Chen, Xiaobin  and
      Meurers, Detmar",
    editor = "Brunato, Dominique  and
      Dell{'}Orletta, Felice  and
      Venturi, Giulia  and
      Fran{\c{c}}ois, Thomas  and
      Blache, Philippe",
    booktitle = "Proceedings of the Workshop on Computational Linguistics for Linguistic Complexity ({CL}4{LC})",
    month = dec,
    year = "2016",
    address = "Osaka, Japan",
    publisher = "The COLING 2016 Organizing Committee",
    url = "https://aclanthology.org/W16-4113/",
    pages = "113--119"
}

@inproceedings{abercrombie-etal-2025-consistency,
    title = "Consistency is Key: Disentangling Label Variation in Natural Language Processing with Intra-Annotator Agreement",
    author = "Abercrombie, Gavin  and
      Dinkar, Tanvi  and
      Cercas Curry, Amanda  and
      Rieser, Verena  and
      Hovy, Dirk",
    editor = "Abercrombie, Gavin  and
      Basile, Valerio  and
      Frenda, Simona  and
      Tonelli, Sara  and
      Dudy, Shiran",
    booktitle = "Proceedings of the The 4th Workshop on Perspectivist Approaches to NLP",
    month = nov,
    year = "2025",
    address = "Suzhou, China",
    publisher = "Association for Computational Linguistics",
    url = "https://aclanthology.org/2025.nlperspectives-1.6/",
    doi = "10.18653/v1/2025.nlperspectives-1.6",
    pages = "63--74",
    ISBN = "979-8-89176-350-0"
}

@inproceedings{glandorf-meurers-2024-towards,
    title = "Towards Fine-Grained Pedagogical Control over {E}nglish Grammar Complexity in Educational Text Generation",
    author = "Glandorf, Dominik  and
      Meurers, Detmar",
    booktitle = "Proc. of the 19th Workshop on Innovative Use of NLP for Building Educational Applications (BEA 2024)",
    month = jun,
    year = "2024",
    url = "https://aclanthology.org/2024.bea-1.24",
    pages = "299--308",
}

@article{okeefe2017english,
  title={{The English Grammar Profile of learner competence: Methodology and key findings}},
  author={O’Keeffe, Anne and Mark, Geraldine},
  journal={International Journal of Corpus Linguistics},
  volume={22},
  number={4},
  pages={457--489},
  year={2017},
  publisher={John Benjamins},
doi={https://doi.org/10.1075/ijcl.14086.oke}
}

@article{mcnamara2010linguistic,
  title={Linguistic features of writing quality},
  author={McNamara, Danielle S and Crossley, Scott A and McCarthy, Philip M},
  journal={Written Communication},
  volume={27},
  number={1},
  pages={57--86},
  year={2010},
  publisher={Sage Publications Sage CA: Los Angeles, CA},
doi={https://doi.org/10.1177/0741088309351547}
}

@book{klebanov2022automated,
  title={Automated essay scoring},
  author={Klebanov, Beata Beigman and Madnani, Nitin},
  year={2022},
  publisher={Springer Nature},
doi={https://doi.org/10.1007/978-3-031-02182-4},
address={n.p.}
}

@article{azar2007grammar,
  title={Grammar-Based Teaching: A Practitioner's Perspective},
  author={Azar, Betty},
  journal={TESL-EJ},
  volume={11},
  number={2},
  year={2007}
}

@article{bulte2012defining,
  title={{Defining and operationalising L2 complexity}},
  author={Bult{\'e}, Bram and Housen, Alex},
  journal={{Dimensions of L2 performance and proficiency: Complexity, accuracy and fluency in SLA}},
  volume={32},
  pages={21},
  year={2012},
doi={https://doi.org/10.1075/lllt.32.02bul}
}

@article{foster1996, title={The Influence of Planning and Task Type on Second Language Performance}, volume={18}, DOI={10.1017/S0272263100015047}, number={3}, journal={Studies in Second Language Acquisition}, author={Foster, Pauline and Skehan, Peter}, year={1996}, pages={299–323}}

@article{lan2019grammaticalcomplexity,
title = {{Grammatical complexity: ‘What Does It Mean’ and ‘So What’ for L2 writing classrooms?}},
journal = {Journal of Second Language Writing},
volume = {46},
pages = {100673},
year = {2019},
issn = {1060-3743},
doi = {https://doi.org/10.1016/j.jslw.2019.100673},
url = {https://www.sciencedirect.com/science/article/pii/S1060374319303649},
author = {Ge Lan and Qiandi Liu and Shelley Staples}
}

@inproceedings{li2024essay,
  title     = {Automated Essay Scoring: Recent Successes and Future Directions},
  author    = {Li, Shengjie and Ng, Vincent},
  booktitle = {Proc. of the Thirty-Third International Joint Conference on
               Artificial Intelligence, {IJCAI-24}},
  pages     = {8114--8122},
  year      = {2024},
  month     = {8},
  note      = {{Survey Track}},
  doi       = {10.24963/ijcai.2024/897},
  url       = {https://doi.org/10.24963/ijcai.2024/897},
}

@incollection{gamon,
  author      = "Gamon, M. and Chodorow, M. and Leacock, C. and Tetreault, J.",
  title       = "Grammatical Error Detection in Automatic Essay Scoring and Feedback",
  editor      = "Shermis, {M.D.} and Burstein, {J.C.}",
  booktitle   = "Handbook of Automated Essay Evaluation",
  publisher   = "Routledge",
  address     = "New York",
  year        = 2013,
  pages       = "251-266",
  chapter     = 15,
  doi={10.4324/9780203122761}
}

@inproceedings{ide2025make,
  title={{How to Make the Most of LLMs’ Grammatical Knowledge for Acceptability Judgments}},
  author={Ide, Yusuke and Nishida, Yuto and Vasselli, Justin and others},
  booktitle={Proc. of the 2025 Conference of the Nations of the Americas Chapter of the Association for Computational Linguistics: Human Language Technologies (Volume 1: Long Papers)},
  pages={7416--7432},
  year={2025},
doi={10.18653/v1/2025.naacl-long.380}
}

@inproceedings{sravanthi-etal-2024-pub,
    title = "{PUB}: A Pragmatics Understanding Benchmark for Assessing {LLM}s' Pragmatics Capabilities",
    author = "Sravanthi, Settaluri  and
      Doshi, Meet  and
      Tankala, Pavan  and
      others",
    booktitle = "Findings of the Association for Computational Linguistics: ACL 2024",
    month = aug,
    year = "2024",
    url = "https://aclanthology.org/2024.findings-acl.719/",
    doi = "10.18653/v1/2024.findings-acl.719",
    pages = "12075--12097",
}

@article{chau2023comparison,
  title={{A comparison of automated and manual analyses of syntactic complexity in L2 English writing}},
  author={Ch{\^a}u, Quang Hong and Bult{\'e}, Bram},
  journal={International Journal of Corpus Linguistics},
  volume={28},
  number={2},
  pages={232--262},
  year={2023},
  publisher={John Benjamins Publishing Company Amsterdam/Philadelphia},
doi={https://doi.org/10.1075/ijcl.20181.cha}
}

@inproceedings{kennedy-2025-evidence,
    title = "Evidence of Generative Syntax in {LLM}s",
    author = "Kennedy, Mary",
    editor = "Boleda, Gemma  and
      Roth, Michael",
    booktitle = "Proc. of the 29th Conference on Computational Natural Language Learning",
    month = jul,
    year = "2025",
    url = "https://aclanthology.org/2025.conll-1.25/",
    doi = "10.18653/v1/2025.conll-1.25",
    pages = "377--396",
    ISBN = "979-8-89176-271-8",
}

@article{kyle2018measuring,
  title={{Measuring syntactic complexity in L2 writing using fine-grained clausal and phrasal indices}},
  author={Kyle, Kristopher and Crossley, Scott A},
  journal={The Modern Language Journal},
  volume={102},
  number={2},
  pages={333--349},
  year={2018},
  publisher={Wiley Online Library},
doi={https://doi.org/10.1111/modl.12468}
}

@inproceedings{yannakoudakis2011,
    title = "A New Dataset and Method for Automatically Grading {ESOL} Texts",
    author = "Yannakoudakis, Helen  and
      Briscoe, Ted  and
      Medlock, Ben",
    booktitle = "Proc. of the 49th Annual Meeting of the Association for Computational Linguistics: Human Language Technologies",
    month = jun,
    year = "2011",
    url = "https://aclanthology.org/P11-1019",
    pages = "180--189",
}

@book{wolfe1998second,
  title={Second Language Development in Writing: Measures of Fluency, Accuracy, \& Complexity},
  author={Wolfe-Quintero, Kathryn Elizabeth and Inagaki, Shunji and Kim, Hae-Young},
  publisher={University of Hawai'i, Second Language Teaching \& Curriculum Center},
  year={1998},
  address={Honolulu, HI}
}

@article{biber2011,
author = {Biber, Douglas and Gray, Bethany and Poonpon, Kornwipa},
title = {{Should We Use Characteristics of Conversation to Measure Grammatical Complexity in L2 Writing Development?}},
journal = {TESOL Quarterly},
volume = {45},
number = {1},
pages = {5-35},
doi = {https://doi.org/10.5054/tq.2011.244483},
url = {https://onlinelibrary.wiley.com/doi/abs/10.5054/tq.2011.244483},
eprint = {https://onlinelibrary.wiley.com/doi/pdf/10.5054/tq.2011.244483},
year = {2011}
}

@inproceedings{naismith-etal-2023-automated,
    title = "Automated evaluation of written discourse coherence using {GPT}-4",
    author = "Naismith, Ben  and
      Mulcaire, Phoebe  and
      Burstein, Jill",
    booktitle = "Proc. of the 18th Workshop on Innovative Use of NLP for Building Educational Applications (BEA 2023)",
    month = jul,
    year = "2023",
    url = "https://aclanthology.org/2023.bea-1.32",
    doi = "10.18653/v1/2023.bea-1.32",
    pages = "394--403",
     
}

@article{biber2016predicting,
  title={Predicting patterns of grammatical complexity across language exam task types and proficiency levels},
  author={Biber, Douglas and Gray, Bethany and Staples, Shelley},
  journal={Applied Linguistics},
  volume={37},
  number={5},
  pages={639--668},
  year={2016},
  publisher={Oxford University Press},
doi={https://doi.org/10.1093/applin/amu059}
}

@article{lu2017automated,
  title={{Automated measurement of syntactic complexity in corpus-based L2 writing research and implications for writing assessment}},
  author={Lu, Xiaofei},
  journal={Language Testing},
  volume={34},
  number={4},
  pages={493--511},
  year={2017},
  publisher={SAGE Publications Sage UK: London, England},
doi={https://doi.org/10.1177/0265532217710675}
}

@inproceedings{chen2011computing,
    title = "Computing and Evaluating Syntactic Complexity Features for Automated Scoring of Spontaneous Non-Native Speech",
    author = "Chen, Miao  and
      Zechner, Klaus",
    booktitle = "Proc. of the 49th Annual Meeting of the Association for Computational Linguistics: Human Language Technologies",
    month = jun,
    year = "2011",
    url = "https://aclanthology.org/P11-1073",
    pages = "722--731",
}

@inproceedings{yoon2012assessment,
    title = "Assessment of {ESL} Learners{'} Syntactic Competence Based on Similarity Measures",
    author = "Yoon, Su-Youn  and
      Bhat, Suma",
    booktitle = "Proc. of the 2012 Joint Conference on Empirical Methods in Natural Language Processing and Computational Natural Language Learning",
    month = jul,
    year = "2012",
    url = "https://aclanthology.org/D12-1055",
    pages = "600--608",
}

@inproceedings{ng2014conll,
    title = "The {C}o{NLL}-2014 Shared Task on Grammatical Error Correction",
    author = "Ng, Hwee Tou  and
      Wu, Siew Mei  and
      Briscoe, Ted  and
      others",
    booktitle = "Proc. of the Eighteenth Conference on Computational Natural Language Learning: Shared Task",
    month = jun,
    year = "2014",
    doi = "10.3115/v1/W14-1701",
    pages = "1--14",
}

@inproceedings{bryant2019bea,
    title = "The {BEA}-2019 Shared Task on Grammatical Error Correction",
    author = "Bryant, Christopher  and
      Felice, Mariano  and
      Andersen, {\O}istein E.  and
      Briscoe, Ted",
    booktitle = "Proc. of the Fourteenth Workshop on Innovative Use of NLP for Building Educational Applications",
    month = aug,
    year = "2019",
    doi = "10.18653/v1/W19-4406",
    pages = "52--75",
}

@inproceedings{bryant2017automatic,
    title = "Automatic Annotation and Evaluation of Error Types for Grammatical Error Correction",
    author = "Bryant, Christopher  and
      Felice, Mariano  and
      Briscoe, Ted",
    booktitle = "Proc. of the 55th Annual Meeting of the Association for Computational Linguistics (Volume 1: Long Papers)",
    month = jul,
    year = "2017",
    doi = "10.18653/v1/P17-1074",
    pages = "793--805",
}

@article{bryant2023grammatical,
    author = {Bryant, Christopher and Yuan, Zheng and Qorib, Muhammad Reza and others},
    title = "{Grammatical Error Correction: A Survey of the State of the Art}",
    journal = {Computational Linguistics},
    pages = {1-59},
    year = {2023},
doi={https://doi.org/10.1162/coli_a_00478}
}

@inproceedings{banno-etal-2024-gpt,
    title = "Can {GPT}-4 do {L}2 analytic assessment?",
    author = "Bannò, Stefano  and
      Vydana, Hari Krishna  and
      Knill, Kate  and
      Gales, Mark",
    booktitle = "Proc. of the 19th Workshop on Innovative Use of NLP for Building Educational Applications (BEA 2024)",
    month = jun,
    year = "2024",
    url = "https://aclanthology.org/2024.bea-1.14",
    pages = "149--164",
     
}

@article{ehret2023measuring,
  title={Measuring language complexity: Challenges and opportunities},
  author={Ehret, Katharina and Berdicevskis, Aleksandrs and Bentz, Christian and Blumenthal-Dram{\'e}, Alice},
  journal={Linguistics Vanguard},
  volume={9},
  number={s1},
  pages={1--8},
  year={2023},
  publisher={De Gruyter},
doi={https://doi.org/10.1515/lingvan-2022-0133}
}

@article{bulte2025complexity,
  title={Complexity and difficulty in second language acquisition: A theoretical and methodological overview},
  author={Bult{\'e}, Bram and Housen, Alex and Pallotti, Gabriele},
  journal={Language Learning},
  volume={75},
  number={2},
  pages={533--574},
  year={2025},
  publisher={Wiley Online Library},
doi={https://doi.org/10.1111/lang.12669}
}

@article{biber2025grammatical,
  title={Grammatical analysis is required to describe grammatical (and “syntactic”) complexity},
  author={Biber, Douglas and Gray, Bethany and Larsson, Tove and Staples, Shelley},
  journal={Language Learning},
  year={2025},
doi={https://doi.org/10.1111/lang.12683}
}

@inproceedings{berzak-etal-2016-universalfull,
    title = "{U}niversal {D}ependencies for Learner {E}nglish",
    author = "Berzak, Yevgeni  and
      Kenney, Jessica  and
      Spadine, Carolyn  and
      Wang, Jing Xian  and
      Lam, Lucia  and
      Mori, Keiko Sophie  and
      Garza, Sebastian  and
      Katz, Boris",
    editor = "Erk, Katrin  and
      Smith, Noah A.",
    booktitle = "Proc. of the 54th Annual Meeting of the Association for Computational Linguistics (Volume 1: Long Papers)",
    month = aug,
    year = "2016",
    address = "Berlin, Germany",
    publisher = "Association for Computational Linguistics",
    url = "https://aclanthology.org/P16-1070/",
    doi = "10.18653/v1/P16-1070",
    pages = "737--746"
}

@inproceedings{berzak-etal-2016-universal,
    title = "{U}niversal {D}ependencies for Learner {E}nglish",
    author = "Berzak, Yevgeni  and
      Kenney, Jessica  and
      Spadine, Carolyn  and
      others",
    booktitle = "Proc. of the 54th Annual Meeting of the Association for Computational Linguistics (Volume 1: Long Papers)",
    month = aug,
    year = "2016",
    url = "https://aclanthology.org/P16-1070/",
    doi = "10.18653/v1/P16-1070",
    pages = "737--746"
}

@article{kim2021generalizability,
author = {Susie Kim},
title = {{Generalizability of CEFR Criterial Grammatical Features in a Korean EFL Corpus across A1, A2, B1, and B2 Levels}},
journal = {Language Assessment Quarterly},
volume = {18},
number = {3},
pages = {273--295},
year = {2021},
publisher = {Routledge},
doi = {10.1080/15434303.2020.1855647},
URL = {https://doi.org/10.1080/15434303.2020.1855647},
eprint = {https://doi.org/10.1080/15434303.2020.1855647},
}

@inproceedings{cheng-amiri-2025-linguistic,
    title = "Linguistic Blind Spots of Large Language Models",
    author = "Cheng, Jiali  and
      Amiri, Hadi",
    booktitle = "Proc. of the Workshop on Cognitive Modeling and Computational Linguistics",
    month = may,
    year = "2025",
    url = "https://aclanthology.org/2025.cmcl-1.3/",
    doi = "10.18653/v1/2025.cmcl-1.3",
    pages = "1--17",
}

@article{verratti2025nlp,
  title={{NLP-powered quantitative verification of the English Grammar Profile’s structure-level assignment}},
  author={Verratti-Souto, Daniela and Sagirov, Nelly and Chen, Xiaobin},
  journal={Annual Review of Applied Linguistics},
  pages={1--22},
  year={2025},
  publisher={Cambridge University Press},
doi={10.1017/S0267190525100093}
}

@phdthesis{kyle2016measuring,
  title={Measuring syntactic development in {L2} writing: Fine grained indices of syntactic complexity and usage-based indices of syntactic sophistication},
  author={Kyle, Kristopher},
  year={2016},
  school={ScholarWorks@ Georgia State University}
}

@inproceedings{banno-etal-2025-vocabulary,
    title = "Exploiting the {English Vocabulary Profile} for {L2} word-level vocabulary assessment with {LLMs}",
    author = "Bannò, Stefano  and
      Knill, Kate  and
      Gales, Mark",
    booktitle = "Proc. of the 20th Workshop on Innovative Use of NLP for Building Educational Applications (BEA 2025)",
    month = jul,
    year = "2025",
    url = "https://aclanthology.org/anthology-files/pdf/bea/2025.bea-1.45.pdf",
    pages = "632--646",
doi={10.18653/v1/2025.bea-1.45}
}

@inproceedings{omelianchuk-etal-2024-pillars,
    title = "Pillars of Grammatical Error Correction: Comprehensive Inspection Of Contemporary Approaches In The Era of Large Language Models",
    author = "Omelianchuk, Kostiantyn  and
      Liubonko, Andrii  and
      Skurzhanskyi, Oleksandr  and
      others",
    booktitle = "Proc. of the 19th Workshop on Innovative Use of NLP for Building Educational Applications (BEA 2024)",
    month = jun,
    year = "2024",
    url = "https://aclanthology.org/2024.bea-1.3",
    pages = "17--33",
}

@inproceedings{geertzen2013automatic,
  title={Automatic linguistic annotation of large scale {L2} databases: {The EF-Cambridge Open Language Database (EFCAMDAT)}},
  author={Geertzen, Jeroen and Alexopoulou, Theodora and Korhonen, Anna},
  booktitle={Proc. of the 31st Second Language Research Forum},
  pages={240--254},
  year={2013},
  publisher={Cascadilla Proc. Project},
  address={Somerville},
  url={http://www.lingref.com/cpp/slrf/2012/paper3100.pdf}
}

@inproceedings{katinskaia-yangarber-2024-gpt,
    title = "{GPT}-3.5 for Grammatical Error Correction",
    author = "Katinskaia, Anisia  and
      Yangarber, Roman",
    booktitle = "Proc. of the 2024 Joint International Conference on Computational Linguistics, Language Resources and Evaluation (LREC-COLING 2024)",
    month = may,
    year = "2024",
    url = "https://aclanthology.org/2024.lrec-main.692",
    pages = "7831--7843",
}

@inproceedings{song-etal-2024-gee,
    title = "{GEE}! Grammar Error Explanation with Large Language Models",
    author = "Song, Yixiao  and
      Krishna, Kalpesh  and
      Bhatt, Rajesh  and
      others",
    booktitle = "Findings of the Association for Computational Linguistics: NAACL 2024",
    month = jun,
    year = "2024",
    url = "https://aclanthology.org/2024.findings-naacl.49",
    doi = "10.18653/v1/2024.findings-naacl.49",
    pages = "754--781",
}

@article{yannakoudakis2018developing,
  title={Developing an automated writing placement system for {ESL} learners},
  author={Yannakoudakis, Helen and Andersen, {\O}istein E. and Geranpayeh, Ardeshir and others},
  journal={Applied Measurement in Education},
  volume={31},
  number={3},
  pages={251--267},
  year={2018},
  publisher={Taylor \& Francis},
  doi={10.1080/08957347.2018.1464447}
}

@inproceedings{omelianchuk-etal-2020-gector,
    title = "{GECT}o{R} {--} Grammatical Error Correction: Tag, Not Rewrite",
    author = "Omelianchuk, Kostiantyn  and
      Atrasevych, Vitaliy  and
      Chernodub, Artem  and
      Skurzhanskyi, Oleksandr",
    booktitle = "Proc. of the Fifteenth Workshop on Innovative Use of NLP for Building Educational Applications",
    month = jul,
    year = "2020",

    url = "https://aclanthology.org/2020.bea-1.16",
    doi = "10.18653/v1/2020.bea-1.16",
    pages = "163--170",
}

@book{richards2008moving,
  title={Moving beyond the plateau},
  author={Richards, Jack Croft},
  year={2008},
  publisher={Cambridge University Press New York, NY}
}

@article{staples2016academic,
author = {Shelley Staples and Jesse Egbert and Douglas Biber and Bethany Gray},
title ={Academic Writing Development at the University Level: Phrasal and Clausal Complexity Across Level of Study, Discipline, and Genre},

journal = {Written Communication},
volume = {33},
number = {2},
pages = {149-183},
year = {2016},
doi = {10.1177/0741088316631527},

URL = { 
    
        https://doi.org/10.1177/0741088316631527
    
    

},
eprint = { 
    
        https://doi.org/10.1177/0741088316631527
    
    

}
}

@article{lu2010automatic,
  title={Automatic analysis of syntactic complexity in second language writing},
  author={Lu, Xiaofei},
  journal={International journal of corpus linguistics},
  volume={15},
  number={4},
  pages={474--496},
  year={2010},
  publisher={John Benjamins},
doi={doi={https://doi.org/10.1075/ijcl.15.4.02lu}}
}

@inproceedings{
he2021deberta,
title={{DeBERTA: Decoding-enhanced BERT with disentangled attention}},
author={Pengcheng He and Xiaodong Liu and Jianfeng Gao and Weizhu Chen},
booktitle={International Conference on Learning Representations},
year={2021},
url={https://openreview.net/forum?id=XPZIaotutsD}
}

@inproceedings{weissweiler-etal-2022-better,
    title = "The better your Syntax, the better your Semantics? Probing Pretrained Language Models for the {E}nglish Comparative Correlative",
    author = {Weissweiler, Leonie  and
      Hofmann, Valentin  and
      K{\"o}ksal, Abdullatif  and
      Sch{\"u}tze, Hinrich},
    booktitle = "Proc. of the 2022 Conference on Empirical Methods in Natural Language Processing",
    month = dec,
    year = "2022",
    url = "https://aclanthology.org/2022.emnlp-main.746/",
    doi = "10.18653/v1/2022.emnlp-main.746",
    pages = "10859--10882",
}

@incollection{chujo2015corpus,
  title={{A corpus and grammatical browsing system for remedial EFL learners}},
  author={Chujo, Kiyomi and Oghigian, Kathryn and Akasegawa, Shiro},
  booktitle={Multiple Affordances of Language Corpora for Data-driven Learning},
  pages={109--128},
  year={2015},
  publisher={John Benjamins Publishing Company},
address={Amsterdam},
doi={
https://doi.org/10.1075/scl.69.06chu}
}

@article{sagirov2025polke,
  title={{POLKE: A system for comprehensively annotating pedagogically oriented grammatical structure use in language production}},
  author={Sagirov, Nelly and Chen, Xiaobin},
  journal={Manuscript submitted for publication to Behavior Research Methods},
  year={2025},
doi={https://doi.org/10.31235/osf.io/zyxw3_v1}
}

@article{ferrucci2004uima,
  title={{UIMA: An architectural approach to unstructured information processing in the corporate research environment}},
  author={Ferrucci, David and Lally, Adam},
  journal={Natural Language Engineering},
  volume={10},
  number={3-4},
  pages={327--348},
  year={2004},
  publisher={Cambridge University Press},
doi={https://doi.org/10.1017/S1351324904003523}
}

@article{kluegl2016uima,
  title={{UIMA Ruta: Rapid development of rule-based information extraction applications}},
  author={Kluegl, Peter and Toepfer, Martin and Beck, Philip-Daniel and others},
  journal={Natural Language Engineering},
  volume={22},
  number={1},
  pages={1--40},
  year={2016},
  publisher={Cambridge University Press},
doi={https://doi.org/10.1017/S1351324914000114}
}

@article{davies2009coca,
  title={The 385+ million word {Corpus of Contemporary American English} (1990--2008+): Design, architecture, and linguistic insights},
  author={Davies, Mark},
  journal={International Journal of Corpus Linguistics},
  volume={14},
  number={2},
  pages={159--190},
  year={2009},
  publisher={John Benjamins},
address={10.1075/ijcl.14.2.02dav}
}

@article{qwen2.5full,
    title   = {Qwen2.5 Technical Report}, 
    author  = {An Yang and Baosong Yang and Beichen Zhang and Binyuan Hui and Bo Zheng and Bowen Yu and Chengyuan Li and Dayiheng Liu and Fei Huang and Haoran Wei and Huan Lin and Jian Yang and Jianhong Tu and Jianwei Zhang and Jianxin Yang and Jiaxi Yang and Jingren Zhou and Junyang Lin and Kai Dang and Keming Lu and Keqin Bao and Kexin Yang and Le Yu and Mei Li and Mingfeng Xue and Pei Zhang and Qin Zhu and Rui Men and Runji Lin and Tianhao Li and Tingyu Xia and Xingzhang Ren and Xuancheng Ren and Yang Fan and Yang Su and Yichang Zhang and Yu Wan and Yuqiong Liu and Zeyu Cui and Zhenru Zhang and Zihan Qiu},
    journal = {arXiv preprint arXiv:2412.15115},
    year    = {2024},
doi={
https://doi.org/10.48550/arXiv.2412.15115}
}

\section*{Supplementary material: Annotation instructions}

\subsection{General rules}

As a general rule for all \emph{can-do} statements, the presence of errors unrelated to a given grammatical construct does not affect its outcome. If the construct is correctly realised despite such errors, it is marked as correct.

Along these lines, spelling errors do not affect whether a grammatical construct is considered successfully realised. For example, in EGP 37 (\emph{Can use `enough' to modify adjectives}), if spellings such as \emph{enagh} or \emph{enought} are encountered, the construct is still considered realised, provided it is used in a morphologically (or syntactically) correct form (e.g., \emph{I am not good enagh}).

Another general rule is that if a grammatical construct is realised both correctly and incorrectly within the same sentence, it is marked as incorrect. For example, in the sentence \emph{I love music because it can make us feeling better or badder about our feelings}, EGP 19 (\emph{Can form irregular comparative adjectives}) is not considered to be correctly realised.

\subsection{Specific rules}

Certain \emph{can-do} statements require additional rules to be specified.

\noindent \textbf{EGP 983} (\emph{Can use `everything' as subject, with a singular verb}): As mentioned above, spelling errors do not affect the evaluation of a grammatical construct, provided that it is used in a morphologically correct form. Therefore, errors such as \emph{everithyng} are not counted, whereas forms like \emph{every thing} are counted, as they constitute morphological errors.

\noindent \textbf{EGP 209} (\emph{Can form negative interrogative clauses}): In this case, all types of negative interrogative clauses are considered, including negative question tags (e.g., \emph{isn't it?}) and interrogative infinitive constructions (e.g., \emph{why not go there?}).

\noindent \textbf{EGP 242} (\emph{Can use a finite subordinate clause with time conjunctions, before or after a main clause}): For this \emph{can-do} statement, both the main clause and any coordinate clauses are considered, but not subordinate clauses. Therefore, sentences such as \emph{It was a really memorable moment for me when I arrived there} are counted, whereas sentences such as \emph{It is very special for me because when I was younger, I was very happy when it was summer} are not.

Constructions such as \emph{Autumn is the season when the leaves starts to fall} are not considered, as they involve relative clauses.

Additionally, given the syntactic nature of this \emph{can-do} statement, tense errors are not taken into account (as they are addressed by other \emph{can-do} statements in the EGP).

\noindent \textbf{EGP 249} (\emph{Can use a finite subordinate clause, before or after a main clause, with conjunctions to introduce conditions}): As in the previous \emph{can-do} statement, both the main clause and any coordinate clauses are considered, but not subordinate clauses.

Additionally, given the syntactic nature of this \emph{can-do} statement, tense and mood errors are not taken into account, as the correctness of indicative, subjunctive, and conditional forms in the protasis and apodosis is evaluated by other \emph{can-do} statements in the EGP.

\section*{Supplementary material: W\&I-EGP data selection process}
\label{appendix_data_selection}

To extract the 3,671 sentence pairs used in our experiments on grammatical construct detection and classification, we applied tailored filters for each selected \emph{can-do} statement to the corrected versions of the sentences. This approach was necessary to reduce the number of sentences that would otherwise be categorised as \emph{no attempt} (i.e., $\omega_3$). At the same time, the filter needs to be broad to intentionally include false positives, allowing us to test both the rule-based system's grammatical and syntactic tagger and the LLMs. For each construct, if the filter yielded 400 or more sentences, we randomly selected 300; otherwise, we retained all the filtered sentences. Below, we describe the filtering criteria used for each of the 12 grammatical constructs included in our analysis.

\subsection{Lexical}

\noindent \textbf{EGP 37} (\emph{Can use `enough' to modify adjectives}): We filtered for sentences containing the word \emph{enough}, without additional constraints.

\noindent \textbf{EGP 266} (\emph{Can use `either ... or' to connect two words, phrases or clauses}): We filtered for sentences containing the word \emph{either}, without additional constraints.

\noindent \textbf{EGP 295} (\emph{Can use `another' to talk about something different}): We filtered for sentences containing the word \emph{another}, without additional constraints.

\noindent \textbf{EGP 367} (\emph{Can use `would' to talk about the future from a point in the past}): We filtered for sentences that satisfy the following conditions:

\begin{itemize}
\item [a.] The lemma \emph{would} is followed by an infinitive verb within a window of 5 tokens (to account for possible negations and adverbs);
\item [b.] The sentence contains at least one verb in the past tense.
\end{itemize}

\noindent \textbf{EGP 598} (\emph{Can use `used to' to talk about repeated actions or states in the past that are no longer true}): We filtered for sentences containing \emph{used} and \emph{didn't use}, without additional constraints.

\noindent \textbf{EGP 983} (\emph{Can use `everything' as subject, with a singular verb}): We filtered for sentences containing the word \emph{everything}, without additional constraints.

\subsection{Non-lexical}

\noindent \textbf{EGP 19} (\emph{Can form irregular comparative adjectives}): We filtered for sentences containing the words \emph{better}, \emph{worse}, \emph{further}, \emph{farther}, and \emph{elder}. In order to test the ability of the LLM to distinguish between comparatives and superlatives, we also included sentences containing \emph{eldest}, \emph{best}, \emph{worst}, \emph{furthest}, and \emph{farthest}. Additionally, we also included sentences with \emph{more}, \emph{less}, \emph{least}, and \emph{most}.

\noindent \textbf{EGP 209} (\emph{Can form negative interrogative clauses}): First, we filtered for sentences containing the lemma \emph{not}. Then, we randomly selected 125 sentences without question marks, presumably affirmative statements, and 175 sentences containing question marks, which are likely to be questions.

\noindent \textbf{EGP 228} (\emph{Can use a defining relative clause, without a relative pronoun}): We filtered for sentences containing the \texttt{spaCy} universal dependency \texttt{relcl}, which identifies relative clauses.

\noindent \textbf{EGP 242} (\emph{Can use a finite subordinate clause with time conjunctions, before or after a main clause}): We filtered for sentences containing the words \emph{after}, \emph{before}, \emph{when}, \emph{while}, \emph{as}, \emph{once}, \emph{since}, \emph{until}, \emph{till}, \emph{whenever}, \emph{now}, \emph{long}, and \emph{soon}.

\noindent \textbf{EGP 249} (\emph{Can use a finite subordinate clause, before or after a main clause, with conjunctions to introduce conditions}): We filtered for sentences containing the words \emph{if}, \emph{unless}, \emph{provided}, \emph{providing}, and \emph{supposing}, and expressions \emph{so long as}, \emph{as long as}, and \emph{in case}.

\noindent \textbf{EGP 708} (\emph{Can use the past simple passive affirmative with a range of pronoun and noun subjects}): We filtered for sentences that satisfy the following conditions:

\begin{itemize}
\item [a.] The sentence contains the \texttt{spaCy} universal dependency \texttt{auxpass}, indicating a passive auxiliary;
\item [b.] The sentence contains the words \emph{was} or \emph{were}.
\end{itemize}

\section*{Supplementary material: Rule-based system for grammatical construct detection and classification}
\label{appendix_rule_based}

Below, we describe the rule-based system used for the 12 selected grammatical constructs investigated in our experiments on grammatical construct detection and classification.

\subsection{Lexical}

\noindent \textbf{EGP 37} (\emph{Can use `enough' to modify adjectives}): The rules checks whether the sentence contains a token that satisfies both of the following conditions:

\begin{itemize}
    \item [a.] The token is \emph{enough};

    \item [b.] The token is preceded by a token that is tagged as an adjective (\texttt{ADJ}).
\end{itemize}

\noindent \textbf{EGP 266} (\emph{Can use `either ... or' to connect two words, phrases or clauses}): The rule checks whether the sentence contains the correlative construction \emph{either ... or} by verifying the following conditions:

\begin{itemize}
\item [a.] The token \emph{either} appears in the sentence;
\item [b.] The token \emph{or} appears after \emph{either}.
\end{itemize} 

\noindent \textbf{EGP 295} (\emph{Can use `another' to talk about something different}): The rule checks whether the sentence contains the token \emph{another}.

\noindent \textbf{EGP 367} (\emph{Can use `would' to talk about the future from a point in the past}): The rule checks whether the sentence satisfies the following conditions:

\begin{itemize}
\item [a.] The lemma \emph{would} is followed by an infinitive verb within a window of 5 tokens (to account for possible negations and adverbs);
\item [b.] The sentence contains at least one verb in the past tense.
\end{itemize}

\noindent \textbf{EGP 598} (\emph{Can use `used to' to talk about repeated actions or states in the past that are no longer true}): The rule checks whether the sentence satisfies one of the following conditions:

\begin{itemize}
\item [a.] The phrase \emph{used to} appears, but is not preceded by the lemma \emph{be} (to exclude cases like \emph{is used to});
\item [b.] The phrase \emph{did not use to} or \emph{didn't use to} appears in sequence.
\end{itemize}

\noindent \textbf{EGP 983} (\emph{Can use `everything' as subject, with a singular verb}): The rule checks whether the sentence contains a token that satisfies the following conditions:

\begin{itemize}
\item [a.] The token is \emph{everything};
\item [b.] The token is tagged as an active subject (\texttt{nsubj}) or a passive subject (\emph{nsubjpass}).
\end{itemize}

\subsection{Non-lexical}

\noindent \textbf{EGP 19} (\emph{Can form irregular comparative adjectives}): The rule checks whether the sentence contains a token that satisfies both of the following conditions:

\begin{itemize}
\item [a.] The token is one of the irregular comparatives: \emph{better}, \emph{worse}, \emph{further}, \emph{farther}, or \emph{elder};
\item [b.] The token is tagged as an adjective (\texttt{ADJ}) and has the comparative tag \texttt{JJR}.
\end{itemize}

\noindent \textbf{EGP 209} (\emph{Can form negative interrogative clauses}): The rule checks whether the sentence satisfies the following conditions:

\begin{itemize}
    \item [a.] It contains the lemma \emph{not};

    \item [b.] It contains a question mark.
\end{itemize}

\noindent \textbf{EGP 228} (\emph{Can use a defining relative clause, without a relative pronoun}): The rule checks whether the sentence satisfies the following conditions:

\begin{itemize}
\item [a.] The sentence contains a relative clause (i.e., the universal dependency tag \texttt{relcl});
\item [b.] When examining the full clause, no relative pronoun (\emph{that}, \emph{who}, \emph{whom}, or \emph{which}) is present.
\end{itemize}

\noindent \textbf{EGP 242} (\emph{Can use a finite subordinate clause with time conjunctions, before or after a main clause}): The rule checks whether the sentence satisfies one of the following conditions:

\begin{itemize}
    \item [a.] It contains multi-word expressions such as \emph{as soon as}, \emph{by the time}, or \emph{as long as};
    \item [b.] It contains single-word subordinating conjunctions like \emph{after}, \emph{before}, \emph{when}, \emph{while}, \emph{since}, \emph{as}, \emph{once}, \emph{until}, \emph{till}, and \emph{whenever} tagged as subordinating conjunctions (\texttt{SCONJ}).
\end{itemize}

\noindent \textbf{EGP 249} (\emph{Can use a finite subordinate clause, before or after a main clause, with conjunctions to introduce conditions}): The rule checks whether the sentence satisfies one of the following conditions:

\begin{itemize}
    \item [a.] It contains multi-word expressions such as \emph{so long as}, \emph{as long as}, and \emph{in case};
    \item [b.] It contains single-word conjunctions such as \emph{if}, \emph{unless}, \emph{provided}, \emph{providing}, or \emph{supposing}.
    
\end{itemize}

\noindent \textbf{EGP 708} (\emph{Can use the past simple passive affirmative with a range of pronoun and noun subjects}): The rule checks whether the sentence contains a token that satisfies the following conditions:

\begin{itemize}
\item [a.] The token is \emph{was} or \emph{were};
\item [b.] The token is tagged as a passive auxiliary (\texttt{auxpass}).
\end{itemize}

\section*{Supplementary material: Additional Results}
\label{supmat_add_res}

\begin{figure*}[htbp!]
    \centering
    \begin{minipage}{0.32\linewidth}
        \centering
        \includegraphics[width=\linewidth]{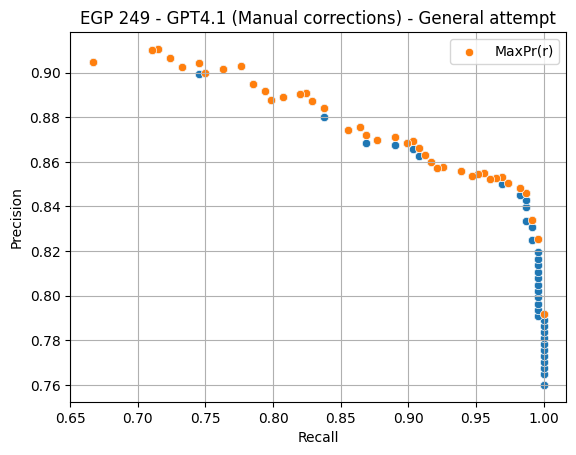}
        
        \label{fig:egp_249att}
    \end{minipage}
    \begin{minipage}{0.32\linewidth}
        \centering
        \includegraphics[width=\linewidth]{egp249_scat_suc.png}
       
        \label{fig:egp_249succ_add}
    \end{minipage}
    \begin{minipage}{0.32\linewidth}
        \centering
        \includegraphics[width=\linewidth]{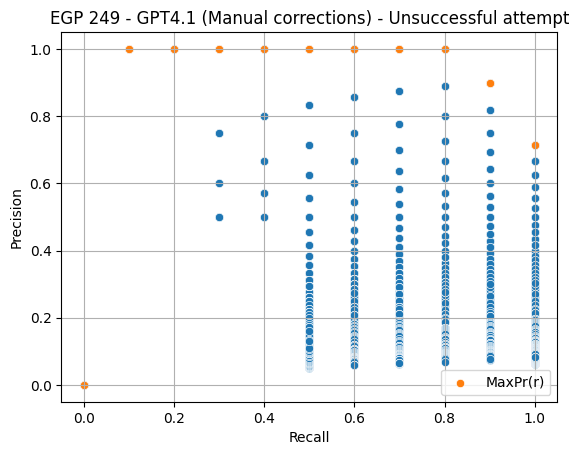}
        
        \label{fig:egp_249unsucc}
    \end{minipage}
    \caption{Precision-Recall scatterplots for EGP 249 (\emph{Can use a finite subordinate clause, before or after a main clause, with conjunctions to introduce conditions}). Orange points represent the upper-bound precision and recall obtained through $\text{MaxPr}(r)$.}
    \label{fig:egp_249_scatter_add}
\end{figure*}

\begin{figure*}[htbp!]
    \centering
    \begin{minipage}{0.32\linewidth}
        \centering
        \includegraphics[width=\linewidth]{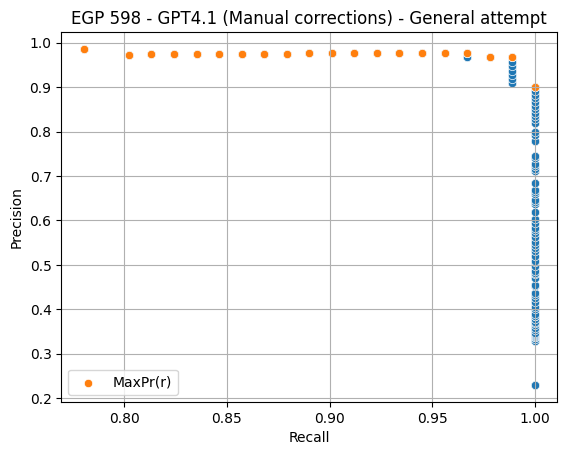}
        
        \label{fig:egp_598att}
    \end{minipage}
    \begin{minipage}{0.32\linewidth}
        \centering
        \includegraphics[width=\linewidth]{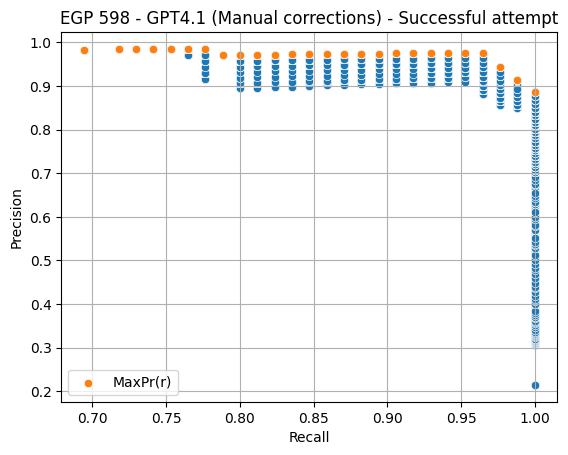}
       
        \label{fig:egp_598succ}
    \end{minipage}
    \begin{minipage}{0.32\linewidth}
        \centering
        \includegraphics[width=\linewidth]{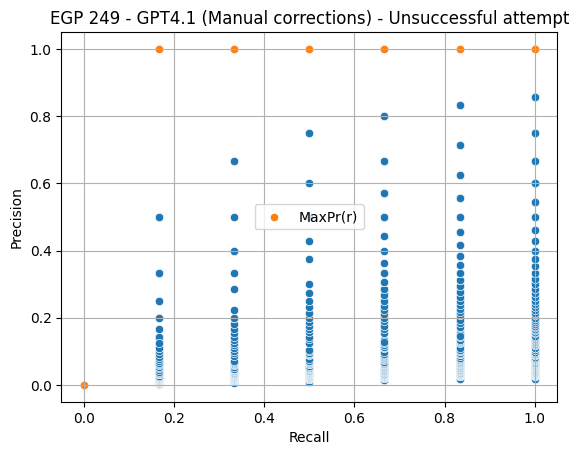}
        
        \label{fig:egp_598unsucc}
    \end{minipage}
    \caption{Precision-Recall scatterplots for EGP 598 (\emph{Can use `used to' to talk about repeated actions or states in the past that are no longer true}). Orange points represent the upper-bound precision and recall obtained through $\text{MaxPr}(r)$.}
    \label{fig:egp_598_scatter_add}
\end{figure*}

Figures \ref{fig:egp_249_scatter_add} and \ref{fig:egp_598_scatter_add} show the Precision-Recall scatterplots for EGP 249 and EGP 598, respectively. These plots provide a visual aid for understanding the procedure used to identify the maximum Precision, $\text{MaxPr}(r)$, values described in Section V-A2 of the main paper.

\begin{figure*}[htbp!]
    \centering
    \begin{minipage}{0.32\linewidth}
        \centering
        \includegraphics[width=\linewidth]{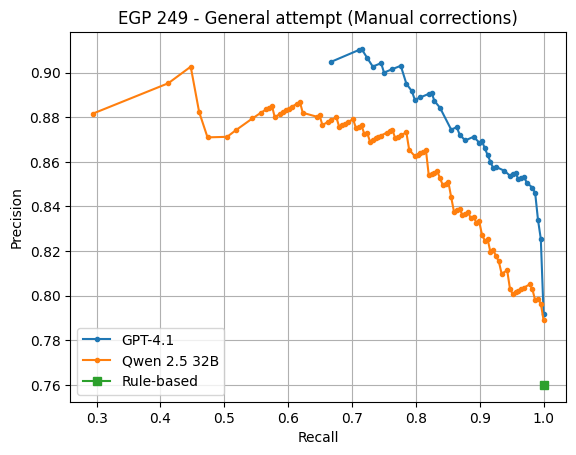}
        
        \label{fig:egp_249att_curve}
    \end{minipage}
    \begin{minipage}{0.32\linewidth}
        \centering
        \includegraphics[width=\linewidth]{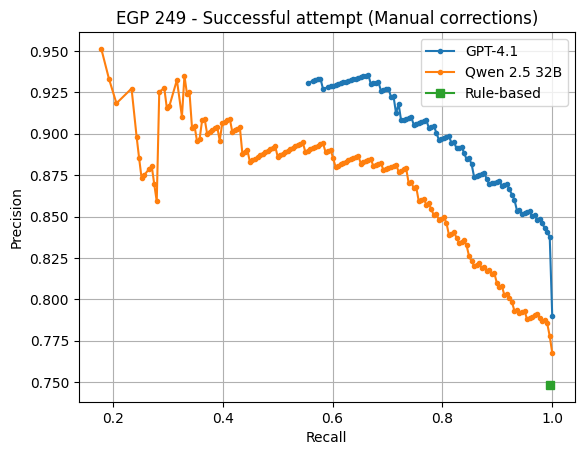}
       
        \label{fig:egp_249succ_curve}
    \end{minipage}
    \begin{minipage}{0.32\linewidth}
        \centering
        \includegraphics[width=\linewidth]{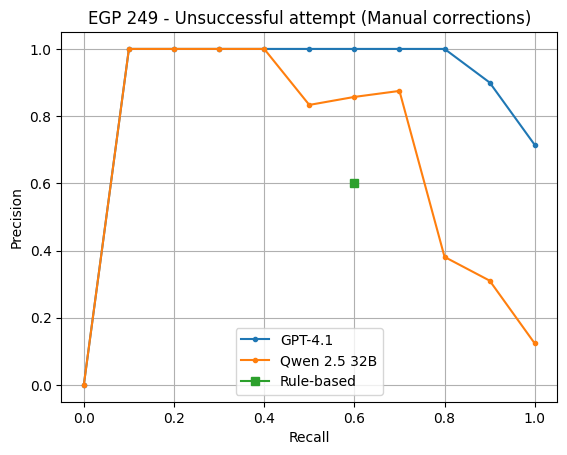}
        
        \label{fig:egp_249unsucc_curve}
    \end{minipage}
    \caption{Precision-Recall curves for EGP 249 (\emph{Can use a finite subordinate clause, before or after a main clause, with conjunctions to introduce conditions}).}
    \label{fig:egp_249}
\end{figure*}

\begin{figure*}[htbp!]
    \centering
    \begin{minipage}{0.32\linewidth}
        \centering
        \includegraphics[width=\linewidth]{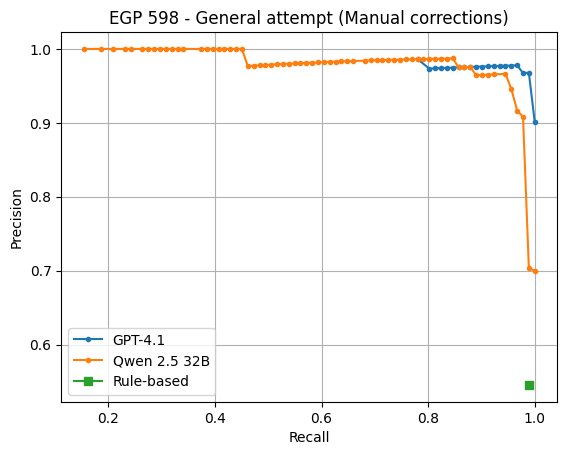}
        
        \label{fig:egp_598att_curve}
    \end{minipage}
    \begin{minipage}{0.32\linewidth}
        \centering
        \includegraphics[width=\linewidth]{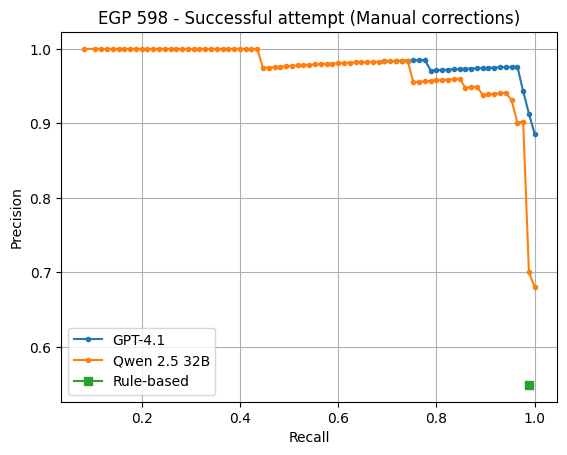}
       
        \label{fig:egp_598succ_curve}
    \end{minipage}
    \begin{minipage}{0.32\linewidth}
        \centering
        \includegraphics[width=\linewidth]{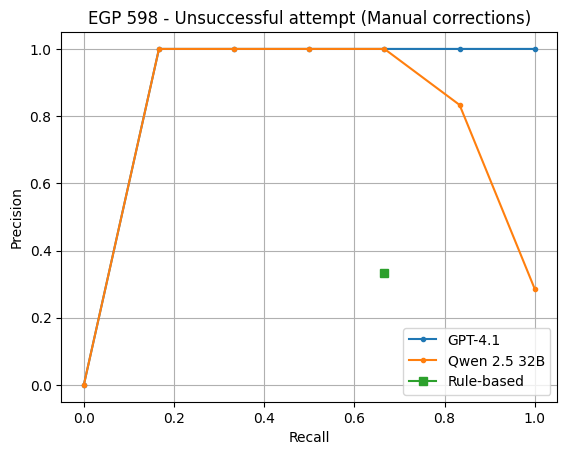}
        
        \label{fig:egp_598unsucc_curve}
    \end{minipage}
    \caption{Precision-Recall curves for EGP 598 (\emph{Can use `used to' to talk about repeated actions or states in the past that are no longer true}).}
    \label{fig:egp_598}
\end{figure*}

Figures \ref{fig:egp_249} and \ref{fig:egp_598} report the Precision-Recall curves for EGP 249 and EGP 598, respectively.

Table \ref{tab:individual_feedback_results_gector} reports the $F_{1}$ scores for each of the 12 selected \emph{can-do} statements using corrections generated through GECToR. 



\begin{table}[ht!]
\centering
\begin{tabular}{c|c|c|c|c|c}
\hline
& \textbf{EGP} & \textbf{Model} & \textbf{Precision} & \textbf{Recall} & \bm{$F_{1}$} \\
\hline
\multirow{24}{*}{\rotatebox{90}{\textbf{Lexical}}} 
& \multirow{4}{*}{37}  & RB       & 97.78  & 88.00  & 92.63  \\ \cmidrule(lr){3-6}
&                         & POLKE    & 97.92   & 94.00   & 95.92   \\ \cmidrule(lr){3-6}
&                         & Qwen     & 32.85  & 90.00  & 48.13  \\ \cmidrule(lr){3-6}
&                         & GPT-4.1  & 93.62  & 88.00  & 90.72  \\ \cmidrule(lr){2-6}

& \multirow{4}{*}{266} & RB       & 100.00  & 100.00  & 100.00  \\ \cmidrule(lr){3-6}
&                        & POLKE    & 100.00  & 37.21  & 54.24 \\ \cmidrule(lr){3-6}
&                        & Qwen     & 100.00  & 100.00  & 100.00  \\ \cmidrule(lr){3-6}
&                        & GPT-4.1  & 100.00  & 100.00  & 100.00  \\ \cmidrule(lr){2-6}

& \multirow{4}{*}{295} & RB       & 60.33  & 100.00  & 75.26  \\ \cmidrule(lr){3-6}
&                        & POLKE    & - & - & - \\ \cmidrule(lr){3-6}
&                        & Qwen     & 61.86  & 99.45  & 76.27  \\ \cmidrule(lr){3-6}
&                        & GPT-4.1  & 85.12  & 79.00  & 81.95  \\ \cmidrule(lr){2-6}

& \multirow{4}{*}{367} & RB       & 15.38  & 84.85  & 26.05  \\ \cmidrule(lr){3-6}
&                        & POLKE    & - & - & - \\ \cmidrule(lr){3-6}
&                        & Qwen     & 68.57  & 72.73  & 70.59  \\ \cmidrule(lr){3-6}
&                        & GPT-4.1  & 75.00  & 81.82  & 78.26  \\ \cmidrule(lr){2-6}

& \multirow{4}{*}{598} & RB       & 54.54  & 98.90  & 70.31  \\ \cmidrule(lr){3-6}
&                        & POLKE    & - & - & - \\ \cmidrule(lr){3-6}
&                        & Qwen     & 96.63  & 94.50  & 95.56  \\ \cmidrule(lr){3-6}
&                        & GPT-4.1  & 96.77  & 98.90  & 97.83  \\ \cmidrule(lr){2-6}

& \multirow{4}{*}{983} & RB       & 93.33  & 97.67  & 95.45   \\ \cmidrule(lr){3-6}
&                        & POLKE    & 94.44   & 79.07   & 86.08  \\ \cmidrule(lr){3-6}
&                        & Qwen     & 95.45  & 97.67  & 96.55  \\ \cmidrule(lr){3-6}
&                        & GPT-4.1  & 96.34   & 91.86   & 94.05   \\ \hline

\multirow{24}{*}{\rotatebox{90}{\textbf{Non-lexical}}}
& \multirow{4}{*}{19}  & RB       & 85.71  & 78.95  & 82.19  \\ \cmidrule(lr){3-6}
&                        & POLKE    & 81.89  & 91.23  & 86.31  \\ \cmidrule(lr){3-6}
&                        & Qwen     & 75.47  & 70.17  & 72.73 \\ \cmidrule(lr){3-6}
&                        & GPT-4.1  & 70.25   & 74.76  & 72.34  \\ \cmidrule(lr){2-6}

& \multirow{4}{*}{209} & RB       & 47.43  & 100.00  & 64.34  \\ \cmidrule(lr){3-6}
&                        & POLKE    & - & - & - \\ \cmidrule(lr){3-6}
&                        & Qwen     & 79.12  & 86.75  & 82.76  \\ \cmidrule(lr){3-6}
&                        & GPT-4.1  & 97.22  & 84.34  & 90.32  \\ \cmidrule(lr){2-6}

& \multirow{4}{*}{228} & RB       & 26.92  & 93.33  & 41.79  \\ \cmidrule(lr){3-6}
&                        & POLKE    & 50.59 & 95.56  & 66.15 \\ \cmidrule(lr){3-6}
&                        & Qwen     & 25.69  & 82.22  & 39.15  \\ \cmidrule(lr){3-6}
&                        & GPT-4.1  & 72.73  & 71.11  & 71.91  \\ \cmidrule(lr){2-6}

& \multirow{4}{*}{242} & RB       & 43.62  & 100.00  & 60.75  \\ \cmidrule(lr){3-6}
&                        & POLKE    & - & - & - \\ \cmidrule(lr){3-6}
&                        & Qwen     & 54.63  & 90.77  & 68.21  \\ \cmidrule(lr){3-6}
&                        & GPT-4.1  & 65.22  & 92.31 & 76.43 \\ \cmidrule(lr){2-6}

& \multirow{4}{*}{249} & RB       & 76.00  & 100.00  & 86.36  \\ \cmidrule(lr){3-6}
&                        & POLKE    & - & - & - \\ \cmidrule(lr){3-6}
&                        & Qwen     & 79.65  & 99.56  & 88.50  \\ \cmidrule(lr){3-6}
&                        & GPT-4.1  & 84.59  & 98.68  & 91.09  \\ \cmidrule(lr){2-6}

& \multirow{4}{*}{708} & RB       & 81.29  & 98.51  & 89.08  \\ \cmidrule(lr){3-6}
&                        & POLKE    & 83.67  & 76.21  & 79.77  \\ \cmidrule(lr){3-6}
&                        & Qwen     & 77.51  & 94.79  & 85.28  \\ \cmidrule(lr){3-6}
&                        & GPT-4.1  & 84.16  & 94.79  & 89.16  \\ \hline
\end{tabular}

\captionof{table}{Results for RB, POLKE, Qwen 2.5 32B, and GPT-4.1 for \textbf{general attempts} in terms of Precision, Recall, and $F_{1}$ score (Manual corrections).}
\label{tab:individual_feedback_general}
\end{table}

\begin{table}[ht!]
\centering
\begin{tabular}{c|c|c|c|c|c}
\hline
& \textbf{EGP} & \textbf{Model} & \textbf{Precision} & \textbf{Recall} & \bm{$F_{1}$} \\
\hline
\multirow{24}{*}{\rotatebox{90}{\textbf{Lexical}}} 
& \multirow{4}{*}{37}  & RB       & 97.44 & 88.37  & 92.68  \\ \cmidrule(lr){3-6}
&                         & POLKE    & 97.62  & 95.35  & 96.47  \\ \cmidrule(lr){3-6}
&                         & Qwen     & 33.02   & 81.39   & 46.98    \\ \cmidrule(lr){3-6}
&                         & GPT-4.1  & 92.68  & 88.37  & 90.48  \\ \cmidrule(lr){2-6}

& \multirow{4}{*}{266} & RB       & 100.00  & 97.06  & 98.51  \\ \cmidrule(lr){3-6}
&                        & POLKE    & 100.00  & 41.18  & 58.33  \\ \cmidrule(lr){3-6}
&                        & Qwen     & 100.00  & 100.00  &  100.00 \\ \cmidrule(lr){3-6}
&                        & GPT-4.1  & 100.00  & 97.06  & 98.51  \\ \cmidrule(lr){2-6}

& \multirow{4}{*}{295} & RB       & 61.51  & 100.00  & 76.17  \\ \cmidrule(lr){3-6}
&                        & POLKE    & - & - & - \\ \cmidrule(lr){3-6}
&                        & Qwen     & 62.55  & 99.39  & 76.78  \\ \cmidrule(lr){3-6}
&                        & GPT-4.1  & 84.37  & 82.82  & 83.59  \\ \cmidrule(lr){2-6}

& \multirow{4}{*}{367} & RB       & 15.38  & 88.00  & 26.19  \\ \cmidrule(lr){3-6}
&                        & POLKE    & - & - & - \\ \cmidrule(lr){3-6}
&                        & Qwen     & 88.23  & 60.00  & 71.43  \\ \cmidrule(lr){3-6}
&                        & GPT-4.1  & 94.74  & 72.00  & 81.82  \\ \cmidrule(lr){2-6}

& \multirow{4}{*}{598} & RB       & 54.90  & 98.82  & 70.59  \\ \cmidrule(lr){3-6}
&                        & POLKE    & - & - & - \\ \cmidrule(lr){3-6}
&                        & Qwen     & 93.10  & 95.29  & 94.19  \\ \cmidrule(lr){3-6}
&                        & GPT-4.1  & 97.62  & 96.47  & 97.04  \\ \cmidrule(lr){2-6}

& \multirow{4}{*}{983} & RB       & 91.43  & 95.52  & 93.43  \\ \cmidrule(lr){3-6}
&                        & POLKE    & 91.07  & 76.12  & 82.93  \\ \cmidrule(lr){3-6}
&                        & Qwen     & 95.52  & 95.52  & 95.52  \\ \cmidrule(lr){3-6}
&                        & GPT-4.1  & 95.45  & 94.03  & 94.74  \\ \hline

\multirow{24}{*}{\rotatebox{90}{\textbf{Non-lexical}}}
& \multirow{4}{*}{19}  & RB       & 85.26  & 77.14  & 81.00  \\ \cmidrule(lr){3-6}
&                        & POLKE    & 81.36  & 91.43  & 86.10  \\ \cmidrule(lr){3-6}
&                        & Qwen     & 75.26  & 69.52  & 72.28  \\ \cmidrule(lr){3-6}
&                        & GPT-4.1  & 68.38  & 76.19  & 72.07  \\ \cmidrule(lr){2-6}

& \multirow{4}{*}{209} & RB       & 37.93  & 98.21  & 54.73  \\ \cmidrule(lr){3-6}
&                        & POLKE    & - & - & - \\ \cmidrule(lr){3-6}
&                        & Qwen     & 67.16  & 80.36  & 73.17  \\ \cmidrule(lr){3-6}
&                        & GPT-4.1  & 85.71  & 85.71  & 85.71  \\ \cmidrule(lr){2-6}

& \multirow{4}{*}{228} & RB       & 28.57  & 95.24  & 43.96  \\ \cmidrule(lr){3-6}
&                        & POLKE    & 53.95  & 97.62  & 69.49  \\ \cmidrule(lr){3-6}
&                        & Qwen     & 25.18  & 83.33  & 38.67  \\ \cmidrule(lr){3-6}
&                        & GPT-4.1  & 76.92  & 71.43  & 74.07  \\ \cmidrule(lr){2-6}

& \multirow{4}{*}{242} & RB       & 42.25  & 100.00  & 59.41  \\ \cmidrule(lr){3-6}
&                        & POLKE    & - & - & - \\ \cmidrule(lr){3-6}
&                        & Qwen     & 52.34  & 93.33  & 67.07 \\ \cmidrule(lr){3-6}
&                        & GPT-4.1  & 64.37  & 93.33  & 76.19  \\ \cmidrule(lr){2-6}

& \multirow{4}{*}{249} & RB       & 74.83  & 99.54  & 85.43  \\ \cmidrule(lr){3-6}
&                        & POLKE    & - & - & - \\ \cmidrule(lr){3-6}
&                        & Qwen     & 78.54  & 99.08  & 87.63  \\ \cmidrule(lr){3-6}
&                        & GPT-4.1  & 83.78  & 99.54  & 90.98  \\ \cmidrule(lr){2-6}

& \multirow{4}{*}{708} & RB       & 77.90  & 97.65  & 86.67  \\ \cmidrule(lr){3-6}
&                        & POLKE    & 80.63  & 72.30  & 76.24  \\ \cmidrule(lr){3-6}
&                        & Qwen     & 75.39  & 90.61  & 82.30  \\ \cmidrule(lr){3-6}
&                        & GPT-4.1  & 79.60 & 93.43  & 85.96 \\ \hline
\end{tabular}

\captionof{table}{Results for RB, POLKE, Qwen 2.5 32B, and GPT-4.1 for \textbf{successful attempts} in terms of Precision, Recall, and $F_{1}$ score (Manual corrections).}
\label{tab:individual_feedback_succ}
\end{table}

\begin{table}[ht!]
\centering
\begin{tabular}{c|c|c|c|c|c}
\hline
& \textbf{EGP} & \textbf{Model} & \textbf{Precision} & \textbf{Recall} & \bm{$F_{1}$} \\
\hline
\multirow{24}{*}{\rotatebox{90}{\textbf{Lexical}}} 
& \multirow{4}{*}{37}  & RB       & 100.00  & 85.71  & 92.31  \\ \cmidrule(lr){3-6}
&                         & POLKE    & 100.00   & 85.71   & 92.31   \\ \cmidrule(lr){3-6}
&                         & Qwen     & 100.00  & 28.57  & 44.44  \\ \cmidrule(lr){3-6}
&                         & GPT-4.1  & 62.50  & 71.43  & 66.67  \\ \cmidrule(lr){2-6}

& \multirow{4}{*}{266} & RB       & 90.00  & 100.00  & 94.74  \\ \cmidrule(lr){3-6}
&                        & POLKE    & 50.00  & 11.11  & 18.18  \\ \cmidrule(lr){3-6}
&                        & Qwen     & 100.00  & 100.00  & 100.00 \\ \cmidrule(lr){3-6}
&                        & GPT-4.1  & 90.00  & 100.00  & 94.74  \\ \cmidrule(lr){2-6}

& \multirow{4}{*}{295} & RB       & 42.86  & 83.33  & 56.60  \\ \cmidrule(lr){3-6}
&                        & POLKE    & - & - & - \\ \cmidrule(lr){3-6}
&                        & Qwen     & 60.00  & 66.67  & 63.16  \\ \cmidrule(lr){3-6}
&                        & GPT-4.1  & 59.09  & 72.22  & 65.00  \\ \cmidrule(lr){2-6}

& \multirow{4}{*}{367} & RB       & 15.38  & 75.00  & 25.53  \\ \cmidrule(lr){3-6}
&                        & POLKE    & - & - & - \\ \cmidrule(lr){3-6}
&                        & Qwen     & 100.00  & 50.00  & 66.67  \\ \cmidrule(lr){3-6}
&                        & GPT-4.1  & 75.00  & 75.00  & 75.00  \\ \cmidrule(lr){2-6}

& \multirow{4}{*}{598} & RB       & 33.33  & 66.67  & 44.44  \\ \cmidrule(lr){3-6}
&                        & POLKE    & - & - & - \\ \cmidrule(lr){3-6}
&                        & Qwen     & 83.33  & 83.33  & 83.33  \\ \cmidrule(lr){3-6}
&                        & GPT-4.1  & 100.00  & 100.00   & 100.00  \\ \cmidrule(lr){2-6}

& \multirow{4}{*}{983} & RB       & 75.00  & 78.95  & 76.92  \\ \cmidrule(lr){3-6}
&                        & POLKE    & 81.25  & 68.42  & 74.29  \\ \cmidrule(lr){3-6}
&                        & Qwen     & 84.21  & 84.21  & 84.21  \\ \cmidrule(lr){3-6}
&                        & GPT-4.1  & 100.00  & 78.95  & 88.23  \\ \hline

\multirow{24}{*}{\rotatebox{90}{\textbf{Non-lexical}}}
& \multirow{4}{*}{19}  & RB       & 70.00  & 77.78  & 73.68  \\ \cmidrule(lr){3-6}
&                        & POLKE    & 55.56  & 55.56  & 55.56  \\ \cmidrule(lr){3-6}
&                        & Qwen     & 50.00  & 66.67  & 57.14  \\ \cmidrule(lr){3-6}
&                        & GPT-4.1  & 50.00  & 77.78  & 60.87  \\ \cmidrule(lr){2-6}

& \multirow{4}{*}{209} & RB       & 30.00  & 33.33  & 31.58 \\ \cmidrule(lr){3-6}
&                        & POLKE    & - & - & - \\ \cmidrule(lr){3-6}
&                        & Qwen     & 64.71  & 40.74  & 50.00  \\ \cmidrule(lr){3-6}
&                        & GPT-4.1  & 87.50  & 51.85  & 65.12  \\ \cmidrule(lr){2-6}

& \multirow{4}{*}{228} & RB       & 0.00  & 0.00  & 0.00  \\ \cmidrule(lr){3-6}
&                        & POLKE    & 0.00  & 0.00  & 0.00 \\ \cmidrule(lr){3-6}
&                        & Qwen     & 20.00  & 33.33  & 25.00  \\ \cmidrule(lr){3-6}
&                        & GPT-4.1  & 33.33  & 33.33  & 33.33  \\ \cmidrule(lr){2-6}

& \multirow{4}{*}{242} & RB       & 42.86  & 60.00  & 50.00  \\ \cmidrule(lr){3-6}
&                        & POLKE    & - & - & - \\ \cmidrule(lr){3-6}
&                        & Qwen     & 100.00  & 40.00  & 57.14  \\ \cmidrule(lr){3-6}
&                        & GPT-4.1  & 100.00  & 60.00  & 75.00  \\ \cmidrule(lr){2-6}

& \multirow{4}{*}{249} & RB       & 60.00  & 60.00  & 60.00  \\ \cmidrule(lr){3-6}
&                        & POLKE    & - & - & - \\ \cmidrule(lr){3-6}
&                        & Qwen     & 87.50  & 70.00  & 77.78  \\ \cmidrule(lr){3-6}
&                        & GPT-4.1  & 90.00  & 90.00  & 90.00  \\ \cmidrule(lr){2-6}

& \multirow{4}{*}{708} & RB       & 71.19  & 75.00  & 73.04 \\ \cmidrule(lr){3-6}
&                        & POLKE    & 74.07  & 71.43  & 72.73  \\ \cmidrule(lr){3-6}
&                        & Qwen     & 68.08  & 57.14  & 62.14  \\ \cmidrule(lr){3-6}
&                        & GPT-4.1  & 80.95  & 60.71  & 69.39  \\ \hline
\end{tabular}

\captionof{table}{Results for RB, POLKE, Qwen 2.5 32B, and GPT-4.1 for \textbf{unsuccessful attempts} in terms of Precision, Recall, and $F_{1}$ score (Manual corrections).}
\label{tab:individual_feedback_unsucc}
\end{table}


\begin{table}[ht!]

\centering

\begin{tabular}{c|c|c|c|c|c}
\hline
& \textbf{EGP} & \textbf{Model} & \textbf{General} & \textbf{Successful} & \textbf{Unsuccessful} \\
\hline
\multirow{24}{*}{\rotatebox{90}{\textbf{Lexical}}} & \multirow{4}{*}{37}  & RB & 85.39  & 92.68    & 0.00    \\ \cmidrule(lr){3-6}
& & POLKE & \textbf{88.17} & \textbf{95.35} & 0.00 \\ \cmidrule(lr){3-6}
                    &  & Qwen & 36.02   & 47.74   & \textbf{44.44}     \\ 
                     \cmidrule(lr){3-6}
& & GPT-4.1 & 85.71  & 92.86 & 18.18  \\ \cmidrule(lr){2-6}

& \multirow{4}{*}{266}  & RB & 88.31   & 98.51      & 20.00     \\ \cmidrule(lr){3-6}
& & POLKE & 59.02 & 61.22 & 0.00 \\ \cmidrule(lr){3-6}
                    &  & Qwen & \textbf{90.00}     & \textbf{100.00}     & \textbf{53.33}   \\  
                    \cmidrule(lr){3-6}
                    & & GPT-4.1 & \textbf{90.00}  & 98.51  & 36.36 \\ \cmidrule(lr){2-6}

& \multirow{4}{*}{295}  & RB & 73.59   & 75.29     & 37.84   \\ \cmidrule(lr){3-6}
& & POLKE & - & - & - \\ \cmidrule(lr){3-6}
                    &  & Qwen & 75.63      & 76.78      & 52.83  \\  
                    \cmidrule(lr){3-6}
                    & & GPT-4.1 & \textbf{79.13} & \textbf{81.70} & \textbf{60.46}  \\ \cmidrule(lr){2-6}

& \multirow{4}{*}{367}  & RB & 24.87  & 26.25   & 18.18   \\ \cmidrule(lr){3-6}
& & POLKE & - & - & - \\ \cmidrule(lr){3-6}
                    &  & Qwen & 65.45   & 73.91   & 54.54      \\ 
                    \cmidrule(lr){3-6}
                    & & GPT-4.1 & \textbf{71.87}  & \textbf{80.00}  & \textbf{57.14}  \\ \cmidrule(lr){2-6}

& \multirow{4}{*}{598}  & RB & 67.47   & 67.24  & 47.06   \\ \cmidrule(lr){3-6}
& & POLKE & - & - & - \\ \cmidrule(lr){3-6}
                    &  & Qwen & 89.65    & 92.22    & 66.67 \\ 
                   \cmidrule(lr){3-6}
                   & & GPT-4.1 & \textbf{91.21}  & \textbf{94.80}  & \textbf{100.00} \\ \cmidrule(lr){2-6}

& \multirow{4}{*}{983}  & RB & 89.41     & 91.97    &  54.54  \\ \cmidrule(lr){3-6}
& & POLKE & 80.00 & 81.60 & 53.33 \\ \cmidrule(lr){3-6}
                    &  & Qwen & \textbf{90.59}    & \textbf{94.81}     &  \textbf{66.67}   \\ \cmidrule(lr){3-6}
                    & & GPT-4.1 & 88.37  & 94.03  & \textbf{66.67}  \\ \hline

\multirow{24}{*}{\rotatebox{90}{\textbf{Non-lexical}}} & \multirow{4}{*}{19}  & RB & 78.70   & 79.21     & 42.86   \\ \cmidrule(lr){3-6}
& & POLKE & \textbf{85.25} & \textbf{85.84} & 11.11 \\ \cmidrule(lr){3-6}
                    &  & Qwen & 70.78      & 71.84   & 42.86  \\  \cmidrule(lr){3-6}
                    & & GPT-4.1 & 72.36  & 71.63  & \textbf{46.15} \\ \cmidrule(lr){2-6}

& \multirow{4}{*}{209}  & RB & 63.64     & 55.84     &  26.67 \\ \cmidrule(lr){3-6}
& & POLKE & - & - & - \\ \cmidrule(lr){3-6}
                    &  & Qwen & 79.04     & 75.97   &  32.26      
                       \\ \cmidrule(lr){3-6}
                       & & GPT-4.1 & \textbf{86.25} & \textbf{86.49}  & \textbf{44.44}  \\ \cmidrule(lr){2-6}

& \multirow{4}{*}{228}  & RB & 45.56   & 45.98    & 0.00   \\ \cmidrule(lr){3-6}
& & POLKE & \textbf{72.88} & \textbf{73.87} & 0.00 \\ \cmidrule(lr){3-6}
                    &  & Qwen &  46.57   &  47.89  &   18.18    \\ \cmidrule(lr){3-6}
                    & & GPT-4.1 & 68.29  & 71.05  & \textbf{40.00}  \\ \cmidrule(lr){2-6}

& \multirow{4}{*}{242}  & RB & 59.15  & 58.82   & 22.22   \\ \cmidrule(lr){3-6}
& & POLKE & - & - & - \\ \cmidrule(lr){3-6}
                    &  & Qwen & 63.83     & 64.37    & 33.33  \\ \cmidrule(lr){3-6}
                    & & GPT-4.1 & \textbf{73.76}  & \textbf{74.07} &  \textbf{50.00} \\ \cmidrule(lr){2-6}

& \multirow{4}{*}{249}  & RB & 85.82    &  85.43   & 28.57   \\ \cmidrule(lr){3-6}
& & POLKE & - & - & - \\ \cmidrule(lr){3-6}
                    &  & Qwen & 87.03     & 87.40     & \textbf{53.33}   \\ \cmidrule(lr){3-6}
                    & & GPT-4.1 & \textbf{89.57}  & \textbf{90.87}  & 52.63 \\ \cmidrule(lr){2-6}

& \multirow{4}{*}{708}  & RB & 82.63      &  84.53    & 47.73     \\ \cmidrule(lr){3-6}
& & POLKE & 73.37 & 74.17  & 39.53 \\ \cmidrule(lr){3-6}
                    &  & Qwen & 80.89     & 82.15     & \textbf{48.70}    \\ \cmidrule(lr){3-6}
                    & & GPT-4.1 & \textbf{83.87}  & \textbf{85.77} & 48.42  \\ \hline

\end{tabular}

\captionof{table}{Results for RB, POLKE, Qwen 2.5 32B, and GPT-4.1 for grammatical construct detection and classification in terms of $F_{1}$ score (GECToR corrections).}
\label{tab:individual_feedback_results_gector}
\end{table}

\begin{table}[ht!]
\centering
\begin{tabular}{c|c|c|c|c|c}
\hline
& \textbf{EGP} & \textbf{Model} & \textbf{Precision} & \textbf{Recall} & \bm{$F_{1}$} \\
\hline
\multirow{24}{*}{\rotatebox{90}{\textbf{Lexical}}} 
& \multirow{4}{*}{37}  & RB       & 97.44 & 76.00  & 85.39  \\ \cmidrule(lr){3-6}
&                         & POLKE    & 95.35  & 82.00  & 88.17  \\ \cmidrule(lr){3-6}
&                         & Qwen     & 23.60  & 76.00  & 36.02  \\ \cmidrule(lr){3-6}
&                         & GPT-4.1  & 95.12  & 78.00  & 85.71  \\ \cmidrule(lr){2-6}

& \multirow{4}{*}{266} & RB       & 100.00  & 79.07  & 88.31  \\ \cmidrule(lr){3-6}
&                        & POLKE    & 100.00  & 41.86  & 59.02  \\ \cmidrule(lr){3-6}
&                        & Qwen     & 97.30  & 83.72  & 90.00  \\ \cmidrule(lr){3-6}
&                        & GPT-4.1  & 97.30  & 83.72  & 90.00  \\ \cmidrule(lr){2-6}

& \multirow{4}{*}{295} & RB       & 60.50  & 93.92  & 73.59 \\ \cmidrule(lr){3-6}
&                        & POLKE    & - & - & - \\ \cmidrule(lr){3-6}
&                        & Qwen     & 61.02  & 99.45  & 75.63  \\ \cmidrule(lr){3-6}
&                        & GPT-4.1  & 90.71  & 70.17  & 79.13  \\ \cmidrule(lr){2-6}

& \multirow{4}{*}{367} & RB       & 15.00  & 72.73  & 24.87  \\ \cmidrule(lr){3-6}
&                        & POLKE    & - & - & - \\ \cmidrule(lr){3-6}
&                        & Qwen     & 81.82  & 54.54  & 65.45  \\ \cmidrule(lr){3-6}
&                        & GPT-4.1  & 74.19  & 69.70  & 71.87 \\ \cmidrule(lr){2-6}

& \multirow{4}{*}{598} & RB       & 53.16  & 92.31  & 67.47  \\ \cmidrule(lr){3-6}
&                        & POLKE    & - & - & - \\ \cmidrule(lr){3-6}
&                        & Qwen     & 93.98  & 85.71  & 89.65  \\ \cmidrule(lr){3-6}
&                        & GPT-4.1  & 91.21 & 91.21  & 91.21  \\ \cmidrule(lr){2-6}

& \multirow{4}{*}{983} & RB       & 90.48  & 88.37  & 89.41  \\ \cmidrule(lr){3-6}
&                        & POLKE    & 89.85  & 72.09  & 80.00  \\ \cmidrule(lr){3-6}
&                        & Qwen     & 91.67  & 89.53  & 90.59  \\ \cmidrule(lr){3-6}
&                        & GPT-4.1  & 88.37  & 88.37  & 88.37  \\ \hline

\multirow{24}{*}{\rotatebox{90}{\textbf{Non-lexical}}}
& \multirow{4}{*}{19}  & RB       & 83.33  & 74.56  & 78.70  \\ \cmidrule(lr){3-6}
&                        & POLKE    & 80.00  & 91.23  & 85.25  \\ \cmidrule(lr){3-6}
&                        & Qwen     & 66.67  & 75.44  & 70.78  \\ \cmidrule(lr){3-6}
&                        & GPT-4.1  & 67.42  & 78.07  & 72.36 \\ \cmidrule(lr){2-6}

& \multirow{4}{*}{209} & RB       & 48.43  & 92.77  & 63.64  \\ \cmidrule(lr){3-6}
&                        & POLKE    & - & - & - \\ \cmidrule(lr){3-6}
&                        & Qwen     & 78.57  & 79.52  & 79.04  \\ \cmidrule(lr){3-6}
&                        & GPT-4.1  & 89.61  & 83.13  & 86.25  \\ \cmidrule(lr){2-6}

& \multirow{4}{*}{228} & RB       & 30.22  & 93.33  & 45.56  \\ \cmidrule(lr){3-6}
&                        & POLKE    & 58.90  & 95.56  & 72.88  \\ \cmidrule(lr){3-6}
&                        & Qwen     & 33.66  & 75.56  & 46.57  \\ \cmidrule(lr){3-6}
&                        & GPT-4.1  & 75.68  & 62.22  & 68.29  \\ \cmidrule(lr){2-6}

& \multirow{4}{*}{242} & RB       & 42.57  & 96.92  & 59.15  \\ \cmidrule(lr){3-6}
&                        & POLKE    & - & - & - \\ \cmidrule(lr){3-6}
&                        & Qwen     & 48.78  & 92.31  & 63.83  \\ \cmidrule(lr){3-6}
&                        & GPT-4.1  & 68.42  & 80.00  & 73.76  \\ \cmidrule(lr){2-6}

& \multirow{4}{*}{249} & RB       & 76.19  & 98.25  & 85.82  \\ \cmidrule(lr){3-6}
&                        & POLKE    & - & - & - \\ \cmidrule(lr){3-6}
&                        & Qwen     & 77.03  & 100.00  & 87.03  \\ \cmidrule(lr){3-6}
&                        & GPT-4.1  & 83.91  & 96.05  & 89.57  \\ \cmidrule(lr){2-6}

& \multirow{4}{*}{708} & RB       & 81.29  & 84.01  & 82.63  \\ \cmidrule(lr){3-6}
&                        & POLKE    & 84.13  & 65.06  & 73.37  \\ \cmidrule(lr){3-6}
&                        & Qwen     & 74.76  & 88.10  & 80.89  \\ \cmidrule(lr){3-6}
&                        & GPT-4.1  & 80.97  & 86.99  & 83.87  \\ \hline
\end{tabular}

\captionof{table}{Results for RB, POLKE, Qwen 2.5 32B, and GPT-4.1 for \textbf{general attempts} in terms of Precision, Recall, and $F_{1}$ score (GECToR corrections).}
\label{tab:individual_feedback_general_gector}
\end{table}

\begin{table}[ht!]
\centering
\begin{tabular}{c|c|c|c|c|c}
\hline
& \textbf{EGP} & \textbf{Model} & \textbf{Precision} & \textbf{Recall} & \bm{$F_{1}$} \\
\hline
\multirow{24}{*}{\rotatebox{90}{\textbf{Lexical}}} 
& \multirow{4}{*}{37}  & RB       & 97.44  & 88.37  & 92.68 \\ \cmidrule(lr){3-6}
&                         & POLKE    & 95.35  & 95.35  & 95.35  \\ \cmidrule(lr){3-6}
&                         & Qwen     & 33.04  & 86.05  & 47.74  \\ \cmidrule(lr){3-6}
&                         & GPT-4.1  & 95.12  & 90.70  & 92.86  \\ \cmidrule(lr){2-6}

& \multirow{4}{*}{266} & RB       & 100.00  & 97.06  & 98.51  \\ \cmidrule(lr){3-6}
&                        & POLKE    & 100.00  & 44.12  & 61.22  \\ \cmidrule(lr){3-6}
&                        & Qwen     & 100.00  & 100.00  & 100.00  \\ \cmidrule(lr){3-6}
&                        & GPT-4.1  & 100.00  & 97.06  & 98.51  \\ \cmidrule(lr){2-6}

& \multirow{4}{*}{295} & RB       & 61.07  & 98.16  & 75.29  \\ \cmidrule(lr){3-6}
&                        & POLKE    & - & - & - \\ \cmidrule(lr){3-6}
&                        & Qwen     & 62.55  & 99.39  & 76.78  \\ \cmidrule(lr){3-6}
&                        & GPT-4.1  & 87.41 & 76.69  & 81.70  \\ \cmidrule(lr){2-6}

& \multirow{4}{*}{367} & RB       & 15.56  & 84.00  & 26.25  \\ \cmidrule(lr){3-6}
&                        & POLKE    & - & - & - \\ \cmidrule(lr){3-6}
&                        & Qwen     & 80.95  & 68.00  & 73.91  \\ \cmidrule(lr){3-6}
&                        & GPT-4.1  & 90.00  & 72.00  & 80.00  \\ \cmidrule(lr){2-6}

& \multirow{4}{*}{598} & RB       & 53.06  & 91.76  & 67.24  \\ \cmidrule(lr){3-6}
&                        & POLKE    & - & - & - \\ \cmidrule(lr){3-6}
&                        & Qwen     & 87.37  & 97.65  & 92.22  \\ \cmidrule(lr){3-6}
&                        & GPT-4.1  & 93.18  & 96.47  & 94.80  \\ \cmidrule(lr){2-6}

& \multirow{4}{*}{983} & RB       & 90.00  & 94.03  & 91.97  \\ \cmidrule(lr){3-6}
&                        & POLKE    & 87.93  & 76.12  & 81.60  \\ \cmidrule(lr){3-6}
&                        & Qwen     & 94.12  & 95.52  & 94.81  \\ \cmidrule(lr){3-6}
&                        & GPT-4.1  & 94.03  & 94.03  & 94.03  \\ \hline

\multirow{24}{*}{\rotatebox{90}{\textbf{Non-lexical}}}
& \multirow{4}{*}{19}  & RB       & 82.47  & 76.19  & 79.21  \\ \cmidrule(lr){3-6}
&                        & POLKE    & 80.16  & 92.38  & 85.84  \\ \cmidrule(lr){3-6}
&                        & Qwen     & 73.27 & 70.48  & 71.84  \\ \cmidrule(lr){3-6}
&                        & GPT-4.1  & 70.00  & 73.33  & 71.63  \\ \cmidrule(lr){2-6}

& \multirow{4}{*}{209} & RB       & 39.01  & 98.21  & 55.84  \\ \cmidrule(lr){3-6}
&                        & POLKE    & - & - & - \\ \cmidrule(lr){3-6}
&                        & Qwen     & 67.12  & 87.50  & 75.97  \\ \cmidrule(lr){3-6}
&                        & GPT-4.1  & 87.27  & 85.71  & 86.49  \\ \cmidrule(lr){2-6}

& \multirow{4}{*}{228} & RB       & 30.30  & 95.24  & 45.98  \\ \cmidrule(lr){3-6}
&                        & POLKE    & 59.42  & 97.62  & 73.87  \\ \cmidrule(lr){3-6}
&                        & Qwen     & 34.00  & 80.95  & 47.89  \\ \cmidrule(lr){3-6}
&                        & GPT-4.1  & 79.41  & 64.29 & 71.05  \\ \cmidrule(lr){2-6}

& \multirow{4}{*}{242} & RB       & 41.67  & 100.00   & 58.82  \\ \cmidrule(lr){3-6}
&                        & POLKE    & - & - & - \\ \cmidrule(lr){3-6}
&                        & Qwen     & 49.12  & 93.33  & 64.37  \\ \cmidrule(lr){3-6}
&                        & GPT-4.1  & 66.67  & 83.33  & 74.07  \\ \cmidrule(lr){2-6}

& \multirow{4}{*}{249} & RB       & 74.83  & 99.54  & 85.43  \\ \cmidrule(lr){3-6}
&                        & POLKE    & - & - & - \\ \cmidrule(lr){3-6}
&                        & Qwen     & 78.47 & 98.62  & 87.40  \\ \cmidrule(lr){3-6}
&                        & GPT-4.1  & 84.58 & 98.16  & 90.87  \\ \cmidrule(lr){2-6}

& \multirow{4}{*}{708} & RB       & 78.86  & 91.08  & 84.53  \\ \cmidrule(lr){3-6}
&                        & POLKE    & 81.46  & 68.07  & 74.17  \\ \cmidrule(lr){3-6}
&                        & Qwen     & 75.79 & 89.67  & 82.15  \\ \cmidrule(lr){3-6}
&                        & GPT-4.1  & 78.29  & 94.84  & 85.77  \\ \hline
\end{tabular}

\captionof{table}{Results for RB, POLKE, Qwen 2.5 32B, and GPT-4.1 for \textbf{successful attempts} in terms of Precision, Recall, and $F_{1}$ score (GECToR corrections).}
\label{tab:individual_feedback_succ_gector}
\end{table}

\begin{table}[ht!]
\centering
\begin{tabular}{c|c|c|c|c|c}
\hline
& \textbf{EGP} & \textbf{Model} & \textbf{Precision} & \textbf{Recall} & \bm{$F_{1}$} \\
\hline
\multirow{24}{*}{\rotatebox{90}{\textbf{Lexical}}} 
& \multirow{4}{*}{37}  & RB       & 0.00  & 0.00  & 0.00  \\ \cmidrule(lr){3-6}
&                         & POLKE    & 0.00  & 0.00  & 0.00  \\ \cmidrule(lr){3-6}
&                         & Qwen     & 100.00  & 28.57  & 44.44  \\ \cmidrule(lr){3-6}
&                         & GPT-4.1  & 13.33  & 28.57  & 18.18  \\ \cmidrule(lr){2-6}

& \multirow{4}{*}{266} & RB       & 100.00  & 11.11  & 20.00  \\ \cmidrule(lr){3-6}
&                        & POLKE    & 0.00  & 0.00  & 0.00  \\ \cmidrule(lr){3-6}
&                        & Qwen     & 66.67  & 44.44  & 53.33  \\ \cmidrule(lr){3-6}
&                        & GPT-4.1  & 100.00  & 22.22  & 36.36  \\ \cmidrule(lr){2-6}

& \multirow{4}{*}{295} & RB       & 36.84  & 38.89  & 37.84  \\ \cmidrule(lr){3-6}
&                        & POLKE    & - & - & - \\ \cmidrule(lr){3-6}
&                        & Qwen     & 40.00  & 77.78  & 52.83  \\ \cmidrule(lr){3-6}
&                        & GPT-4.1  & 52.00  & 72.22  & 60.46 \\ \cmidrule(lr){2-6}

& \multirow{4}{*}{367} & RB       & 12.00  & 37.50  &  18.18 \\ \cmidrule(lr){3-6}
&                        & POLKE    & - & - & - \\ \cmidrule(lr){3-6}
&                        & Qwen     & 42.86  & 75.00  & 54.54  \\ \cmidrule(lr){3-6}
&                        & GPT-4.1  & 66.67  & 50.00  & 57.14  \\ \cmidrule(lr){2-6}

& \multirow{4}{*}{598} & RB       & 36.36  & 66.67  & 47.06  \\ \cmidrule(lr){3-6}
&                        & POLKE    & - & - & - \\ \cmidrule(lr){3-6}
&                        & Qwen     & 100.00  & 50.00  & 66.67  \\ \cmidrule(lr){3-6}
&                        & GPT-4.1  & 100.00  & 100.00  & 100.00  \\ \cmidrule(lr){2-6}

& \multirow{4}{*}{983} & RB       & 64.29  & 47.37  & 54.54  \\ \cmidrule(lr){3-6}
&                        & POLKE    & 72.73  & 42.10  & 53.33  \\ \cmidrule(lr){3-6}
&                        & Qwen     & 78.57  & 57.89  & 66.67  \\ \cmidrule(lr){3-6}
&                        & GPT-4.1  & 90.91  & 52.63  & 66.67  \\ \hline

\multirow{24}{*}{\rotatebox{90}{\textbf{Non-lexical}}}
& \multirow{4}{*}{19}  & RB       & 60.00  & 33.33  & 42.86  \\ \cmidrule(lr){3-6}
&                        & POLKE    & 11.11  & 11.11  & 11.11  \\ \cmidrule(lr){3-6}
&                        & Qwen     & 60.00  & 33.33  & 42.86  \\ \cmidrule(lr){3-6}
&                        & GPT-4.1  & 75.00 & 33.33  & 46.15  \\ \cmidrule(lr){2-6}

& \multirow{4}{*}{209} & RB       & 33.33  & 22.22  & 26.67  \\ \cmidrule(lr){3-6}
&                        & POLKE    & - & - & - \\ \cmidrule(lr){3-6}
&                        & Qwen     & 28.57  & 37.04  & 32.26  \\ \cmidrule(lr){3-6}
&                        & GPT-4.1  & 44.44  & 44.44  & 44.44  \\ \cmidrule(lr){2-6}

& \multirow{4}{*}{228} & RB       & 0.00  & 0.00  & 0.00  \\ \cmidrule(lr){3-6}
&                        & POLKE    & 0.00  & 0.00  & 0.00  \\ \cmidrule(lr){3-6}
&                        & Qwen     & 12.50  & 33.33  & 18.18  \\ \cmidrule(lr){3-6}
&                        & GPT-4.1  & 50.00  & 33.33  & 40.00  \\ \cmidrule(lr){2-6}

& \multirow{4}{*}{242} & RB       & 25.00  & 20.00  & 22.22  \\ \cmidrule(lr){3-6}
&                        & POLKE    & - & - & - \\ \cmidrule(lr){3-6}
&                        & Qwen     & 100.00  & 20.00  & 33.33  \\ \cmidrule(lr){3-6}
&                        & GPT-4.1  & 66.67 & 40.00  & 50.00  \\ \cmidrule(lr){2-6}

& \multirow{4}{*}{249} & RB       & 50.00  & 20.00  & 28.57  \\ \cmidrule(lr){3-6}
&                        & POLKE    & - & - & - \\ \cmidrule(lr){3-6}
&                        & Qwen     & 80.00  & 40.00  & 53.33  \\ \cmidrule(lr){3-6}
&                        & GPT-4.1  & 55.56  & 50.00  & 52.63  \\ \cmidrule(lr){2-6}

& \multirow{4}{*}{708} & RB       & 65.62  & 37.50  & 47.73  \\ \cmidrule(lr){3-6}
&                        & POLKE    & 56.67  & 30.36  & 39.53  \\ \cmidrule(lr){3-6}
&                        & Qwen     & 47.46  & 50.00  & 48.70  \\ \cmidrule(lr){3-6}
&                        & GPT-4.1  & 58.97  & 41.07  & 48.42  \\ \hline
\end{tabular}

\captionof{table}{Results for RB, POLKE, Qwen 2.5 32B, and GPT-4.1 for \textbf{unsuccessful attempts} in terms of Precision, Recall, and $F_{1}$ score (GECToR corrections).}
\label{tab:individual_feedback_unsucc_gector}
\end{table}

\end{document}